\pdfoutput=1

\documentclass[11pt]{article}

\usepackage[final]{acl}

\usepackage{times}
\usepackage{latexsym}

\usepackage[T1]{fontenc}

\usepackage[utf8]{inputenc}

\usepackage{microtype}

\usepackage{inconsolata}

\usepackage{times}
\usepackage{latexsym}

\usepackage{amsmath,amsfonts,bm}

\def\eqref#1{equation~\ref{#1}}

\def\1{\bm{1}}

\DeclareMathAlphabet{\mathsfit}{\encodingdefault}{\sfdefault}{m}{sl}
\SetMathAlphabet{\mathsfit}{bold}{\encodingdefault}{\sfdefault}{bx}{n}

\usepackage{hyperref}
\usepackage{pifont}
\usepackage{array}
\usepackage{xcolor,pifont}
\usepackage{url}
\usepackage{multicol}
\usepackage{multirow}
\usepackage{subcaption}
\usepackage{pdfpages}
\usepackage{inconsolata}
\usepackage{xcolor,pifont,colortbl}
\usepackage[symbol]{footmisc}
\usepackage{microtype}
\usepackage[T1]{fontenc}
\usepackage[utf8]{inputenc}
\usepackage{enumitem}
\usepackage[export]{adjustbox}
\usepackage{booktabs}
\usepackage{amsmath}
\usepackage{mathtools}

\usepackage[T1]{fontenc}

\usepackage[utf8]{inputenc}

\usepackage{microtype}
\usepackage{amssymb}

\newcommand{\Ie}{\textit{I.e.}}
\newcommand{\ie}{\textit{i.e.}}
\newcommand{\Eg}{\textit{E.g.}}
\newcommand{\eg}{\textit{e.g.}}

\title{Tell Me What's Next: Textual Foresight for Generic UI Representations}

\author{Andrea Burns \enspace Kate Saenko \enspace Bryan A. Plummer\\
  Boston University  \\
  \texttt{\{aburns4,saenko,bplum\}@bu.edu}\\
  \url{https://github.com/aburns4/textualforesight}\\}

\begin{document}
\maketitle
\begin{abstract}
Mobile app user interfaces (UIs) are rich with action, text, structure, and image content that can be utilized to learn generic UI representations for tasks like automating user commands, summarizing content, and evaluating the accessibility of user interfaces. Prior work has learned strong visual representations with local or global captioning losses, but fails to retain both granularities.
To combat this,
we propose Textual Foresight, a novel pretraining objective for learning UI screen representations. Textual Foresight generates global text descriptions of \textit{future} UI states given a current UI and local action taken. Our approach requires joint reasoning over elements and entire screens, resulting in improved UI features: on generation tasks, UI agents trained with Textual Foresight outperform state-of-the-art by 2\% with \textbf{28x} fewer images. We train with our newly constructed mobile app dataset, OpenApp, which results in the first public dataset for app UI representation learning. OpenApp enables new baselines, and we find Textual Foresight improves average task performance over them by 5.7\% while having access to \textbf{2x} less data.
\end{abstract}

\section{Introduction}
People use mobile apps every day to browse news articles, shop online, book appointments, and learn from educational platforms~\citep{DOGRUER2011606,appbehavior}. 
AI agents can help to perform these real-life tasks for those who cannot or prefer not to view or touch the app screen (\eg, users who are blind, low-vision, or busy driving)~\citep{screenreader}. To build such AI models, a key question is which modalities should be used to represent the app UI, as it consists of not only the rendered screen, but also metadata, text, and structural features (\ie, the underlying app view hierarchy).

\begin{figure}[t!]
    \centering
    \includegraphics[scale=0.0945]{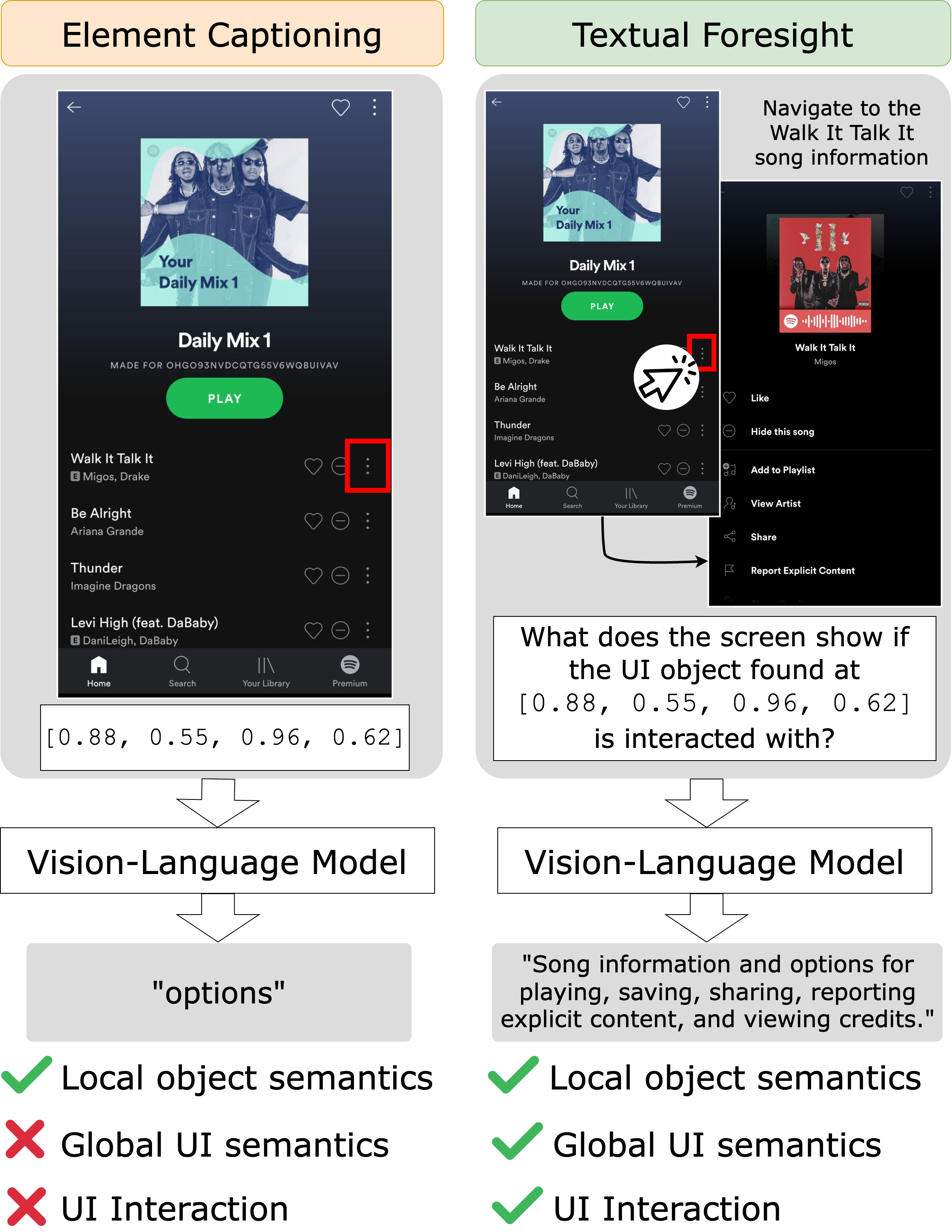}
    \caption{Textual Foresight vs.\ Element Captioning. While both Element Captioning and Textual Foresight pretraining aim to preserve the semantics of individual UI objects, Textual Foresight also requires understanding global UI semantics of the current screen and how an action on the UI will change it, as the objective is to generate the global description of the following screen. We highlight in red the UI object associated with the input bounding box coordinates.}
    \label{fig:motiv}
\end{figure}

Recent work Spotlight learns UI features with only the rendered  screen image~\cite{spotlight}, as the view hierarchy is not always available, and when it is, it often contains generic, noisy, or missing fields~\cite{denoise,burns2022motifvln}. Spotlight proposed UI representation learning via element captioning, and is state-of-the-art on four downstream UI tasks.

While element captioning avoids the disadvantages of other UI modalities, it only enforces local UI understanding.
As shown in Figure~\ref{fig:motiv}(left), this objective trains a model to map an image and bounding box coordinates to an element-level caption like ``options.''
However ``options'' is a limited representation of what this element can do, as it lacks context from the global UI screen or what action it affords.
If we enlarge the visual context to the entire screen, we see that it contains different songs in a streaming application like Spotify. Yet only when seeing the screen that appears upon clicking the ``options'' element, Figure~\ref{fig:motiv}(right), we finally understand that it provides the means to like, hide, or share a particular song.

Our goal is to better balance local element and global screen features, and we find that UI actions can serve as the bridge between them. 
An action performed on a UI informs the semantics of the next UI state. Following this intuition, we propose \textit{Textual Foresight}: a representation learning objective that generates global screen captions of a future UI, given a current UI image and a localized action. This task requires understanding both the local semantics (options icon) and global semantics (a Spotify music playlist) of the current input UI to be able to decode the caption ``song information and options for playing, saving, sharing, reporting explicit content, and viewing credits.'' It also benefits from (state, action) examples, implicitly teaching element affordance.

To study Textual Foresight, we build OpenApp, the first publicly available dataset for representation learning in apps. State-of-the-art Spotlight did not make their pretraining data available, and does not benchmark on a fully open-source evaluate suite, either. We curate OpenApp with multiple element- and screen-level caption sets, which we use to reproduce Spotlight and train other baselines like screen captioning which have never been studied before. We design our framework on top of BLIP-2~\cite{li2023blip2}, making all code publicly available, unlike Spotlight, which also did not open source model code nor checkpoints.

Our experiments show that Textual Foresight is able to better balance the granularity of features learned: it reaches the best average performance for screen summarization and element captioning tasks, which require global and local UI features, respectively. Importantly, Textual Foresight reaches better performance while having 28x less pretraining data than Spotlight, and 2x less than our new baselines. Textual Foresight consistently performs best among our open-source baselines, resulting in a 5.7\% average task performance boost. 

In summary, our contributions include:

\begin{itemize}[nosep,leftmargin=*]
    \item A novel pretraining objective, Textual Foresight, which learns UI representations by describing future UI states given the current screen and a localized action. Textual Foresight outperforms SoTA Spotlight for generation-style tasks with 28x less data.
    \item A new mobile app dataset for UI representation learning, OpenApp, which further annotates and post-processes prior work to make four different pretraining approaches possible. The data is publicly available for download on \href{https://github.com/aburns4/textualforesight}{GitHub}.
    \item The first standardized benchmark for generic UI representations that consists strictly of public datasets for both pretraining and finetuning. We evaluate on element captioning, screen summarization, tappability prediction, and language grounding tasks. All model code and the best checkpoints can be accessed on \href{https://github.com/aburns4/textualforesight}{GitHub}.
\end{itemize}
\section{Related Work}
\begin{figure}[t!]
\centering
\includegraphics[scale=0.188]{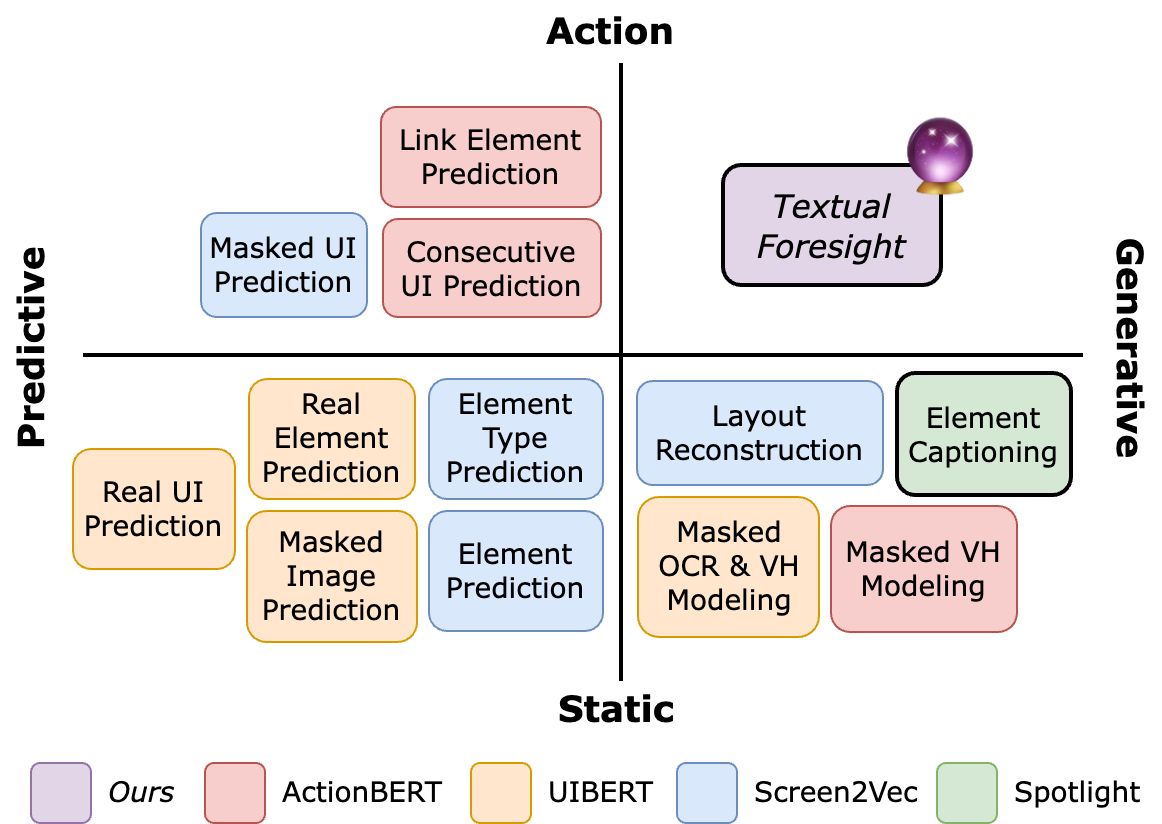}
\caption{Prior Work Comparison. We divide pretraining objectives by loss type (prediction vs. generation) and use of interaction (includes UI actions or only concern static UIs in isolation). We bold Textual Foresight and Element Captioning as they only use the rendered screen to represent the UI.}
\label{fig:losscompare}
\end{figure}

\begin{figure*}[t!]
\centering
\includegraphics[scale=0.1145]{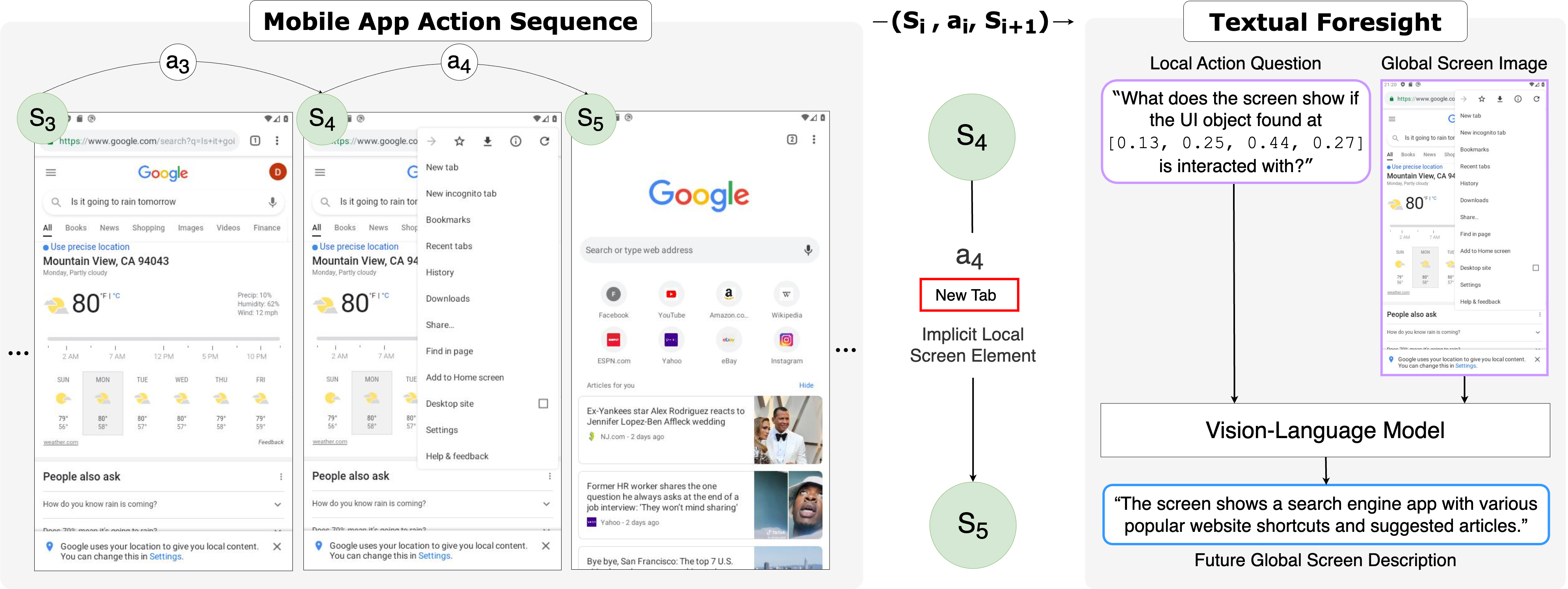}
\caption{Textual Foresight. We illustrate how app states from action sequences are used in (current screen, current
action, next screen) triplets to pretrain a vision-language model for UI representations. We only use the app screen
to represent the UI, and additionally feed in an action question which asks what would we expect to see at the next
state if we interact with a particular UI element. Our model decodes a text description of the following screen, using
action to bridge local element and global screen features. More Textual Foresight examples are in Appendix~\ref{sec:dataexappendix}.}
\label{fig:textfore}
\end{figure*}
While there are several prior methods for learning UI representations, all either use proprietary data and/or evaluate on different tasks, making downstream comparison challenging. Figure~\ref{fig:losscompare} compares Textual Foresight to ActionBERT~\cite{actionbert}, Screen2Vec~\cite{screen2vec}, UIBERT~\cite{bai2021uibert}, and Spotlight~\cite{spotlight}. We compare the type of loss (predictive or generative) and if the loss utilizes action data from the UI. As we see in the upper right quadrant, Textual Foresight is the first generation style loss to incorporate action. Textual Foresight and Spotlight are bolded, as they only input the screen image to represent the UI.

In addition to Textual Foresight and element captioning, global image captioning has been used to learn representations of natural RGB images. \Eg, it is one loss within BLIP-2~\cite{li2023blip2}, which is SoTA on visual question answering, image-text retrieval, and captioning. It has also been used to learn features for vision-only tasks, matching or outperforming SoTA for image classification, object detection, and instance segmentation, while using 10x fewer images during training~\citep{desai2021virtex}. Image captioning has never before been studied as a method to learn UI representations due to a lack of available screen caption data.

We also consider how foresight has been used in prior work. Visual foresight was first introduced to improve robot motion planning~\cite{foresight} and 
has since been incorporated in numerous works in robotics~\citep{robotforesight,pmlr-v100-yen-chen20a}, reinforcement learning~\cite{NEURIPS2018_7ec69dd4,pmlr-v119-nair20a,hierforesight} and vision language navigation~\citep{koh2021pathdreamer,foresightvln}. Differing from Textual Foresight, which predicts language descriptions of future states, these prior works predict raw images~\cite{foresight} or intermediary visual features~\cite{stochasticadv,stochasticvar,foresightvln}. 

Finally, we note that there are several prior works on multimodal UI tasks, datasets, and pretraining approaches in the context of webpage understanding. These works have studied multimodal web agents~\cite{koh2024visualwebarena,yao2022webshop}, multimodal web summarization~\cite{burns-etal-2023-suite}, and web captioning~\cite{wit}.
\section{UI Representation Learning with Textual Foresight}
\label{sec:method}
We aim to learn strong generic UI representations that can be used across many downstream UI tasks. Given a UI screen image $s_t$, the goal of Textual Foresight is to describe what follows from taking action $a_t$ on it. By training a Vision-Language Model (VLM) with Textual Foresight, a single loss can encourage the visual representations to retain both local and global features over the UI screen. 

In Figure~\ref{fig:textfore}, we show how we can learn meaningful features over the input screen image by asking a foresight question. We input a single UI from a longer action sequence, like the Chrome browser state with options opened, and ask what is expected from clicking on ``new tab.'' Visually understanding the new tab element in isolation does not tell us much about the current screen or how interacting with the element would be useful. Yet to be able to describe the \textit{future} UI  as ``a search engine app with various popular website shorted and suggested articles,'' it requires learning a UI representation that captures not only the semantics of ``New Tab,'' but also the global visual context that the input UI contained a search engine result screen.

In Section~\ref{subsec:defn} we define the training loss for Textual Foresight, and then detail model pretraining and finetuning in Section~\ref{subsec:blippre} and~\ref{subsec:blipfine}, respectively.
\subsection{Textual Foresight Definition}
\label{subsec:defn}
Formally, given a current UI screen state $s_t$ and an action performed on it $a_t$, the task of Textual Foresight is to generate a caption $c_{s_{t+1}}$ describing the \textit{next} screen, $s_{t+1}$. 
We train the VLM to decode a foresight caption $c_{s_{t+1}}$ given the prior screen's image $s_t$. To be able to reason about the following UI, we additionally input a question $Q$ which guides the model by asking what is expected after acting upon a particular element:
\begin{align*}
    Q &= \text{``What does the screen show if the UI object} \\
    & \text{found at $[x_1, y_1, x_2, y_2]$ is interacted with?''}
\end{align*}
\noindent 
The $[x_1, y_1, x_2, y_2]$ element bounding box contains the normalized screen coordinates that fall between $[0, 1]$.
We include this bounding box as a part of $Q$, which is ultimately embedded by a language model. This differs from Spotlight, which learns a separate element coordinate embedding. Note that we do not describe an element by its text in $Q$ to ensure the model utilizes the visual context, instead of cheating by only using the element text to infer what might be seen in the future app state.

The model is trained to maximize the probability of the target foresight caption with a cross entropy ($xe$) language modeling loss, similar to many prior captioning approaches~\cite{vinyalsCVPR2015,2020t5,li2023blip2}. Specifically, we minimize the negative log likelihood of the correct word from a vocabulary $V$ at each decoding step $i$. Thus, the Textual Foresight loss can be defined as
\begin{align*}
       L_{foresight} &= L_{xe}(c_{s_{t+1}}, \hat{c}_{s_{t+1}})
\end{align*}
for target caption $c_{s_{t+1}}$ and predicted caption $\hat{c}_{s_{t+1}}$:
\begin{align*}
      c_{s_{t+1}} &=(w_0, w_1, ... w_n) \\
      \hat{c}_{s_{t+1}} &= VLM(Q, s_t)
\end{align*}
where the ground truth caption consists of words $w_i$ and the predicted caption is generated by the VLM with the foresight question $Q$ and the screen state $s_t$ as inputs. Given the target distribution $p$ and the VLM learned distribution $\hat{p}$ over the vocabulary, the cross entropy language modeling loss becomes
\begin{align*}
     L_{xe}(c_{s_{t+1}}, \hat{c}_{s_{t+1}}) &= -p(c_{s_{t+1}})log(\hat{p}(c_{s_{t+1}}))\\
     &= -\sum_{i=0}^{n}\sum_{j=0}^{|V|}p(w_{ij})log(\hat{p}(w_{ij}))\\
     &= -\sum_{i=0}^{n} log(\hat{p}(w_i|w_{<i}))
\end{align*}
The probability distribution $\hat{p}$ over the vocabulary is determined by Softmax outputs from the VLM.

Textual Foresight differs from standard image captioning in two keys ways. First, instead of predicting a caption about the input image $s_t$, we predict a caption about an \textit{unseen} future image $s_{t+1}$. Despite captioning the future screen $s_{t+1}$, we ultimately are refining the features of the input image screen $s_t$; to describe the next UI, the visual representations of the input UI must capture its high-level global semantics \textit{and} the semantics of the action taken on it.

Second, as our task requires a question $Q$ with localized action information, Textual Foresight is in some ways similar to a visual question answering task.
While both Textual Foresight and element captioning require grounded UI understanding, Textual Foresight aims to generate (future) global screen captions. This has the advantage of learning from $(s_t, a_t, s_{t+1})$ samples where $a_t$ corresponds to elements with noisy text or no text at all, which would otherwise be unusable for element captioning. 

\subsection{Pretraining Model}
\label{subsec:blippre}
When learning generic representations, a VLM can first be pretrained with different data and learning objectives than those used to model specific downstream tasks. We apply the BLIP-2 framework~\citep{li2023blip2} for our UI representation learning pretraining and finetuning strategy.

BLIP-2 was originally pretrained in two stages, with the first stage focused on learning to query image representations from a frozen ViT model~\cite{dosovitskiy2020vit}. The query embeddings are learned with an intermediate Transformer, \ie,  Q-Former,~\cite{transformer} with image captioning, image-text contrastive, and image-text matching losses.
The second stage of pretraining continues to train the Q-Former with an image captioning objective while the language model is frozen, adapting the visual queries to useful LLM inputs.

These learned queries are ultimately used as the visual features input to the language model during downstream task finetuning. 
We only pretrain the second stage of BLIP-2 (similar to InstructBLIP~\cite{dai2023instructblip}). In stage two pretraining, we replace the image captioning objective with our Textual Foresight loss. As a result, our representation learning pipeline refines the Q-Former to obtain better visual query embeddings. These improved embeddings serve as our visual representations to the language model when modeling different downstream UI tasks. 

\subsection{Finetuning Model}
\label{subsec:blipfine}

After pretraining the upstream BLIP-2 model with Textual Foresight, we train a different downstream BLIP-2 model for each UI task (\eg, element captioning or tappability prediction). We follow the finetuning procedure as defined in BLIP-2: the ViT model and Q-Former weights are trainable during finetuning, allowing for task-specific representation updates. The LLM (either a FlanT5~\cite{chung2022scaling} encoder-decoder or OPT~\cite{zhang2022opt} decoder-only model) is kept frozen. 

\section{OpenApp Dataset}
\label{sec:pretraindata}
As shown in Figure~\ref{fig:textfore}, Textual Foresight requires mobile app action sequences. In addition to needing data for our new method, other baselines have never been explored due to data limitations (\eg, large scale screen captioning data did not exist) or only studied in a proprietary setting (\eg, element captioning data used in Spotlight).

To curate pretraining data for Textual Foresight and important baselines, we combine and generate new data for existing app datasets MoTIF~\citep{burns2022motifvln}, one snapshot from the longitudinal study by~\citet{fok_longa11y_2022}, and Android in the Wild~\citep[AITW]{rawles2023android}. We refer to the merged data source that we further annotate and post process as OpenApp. The raw OpenApp data consists of app action sequences, with each time step having an action annotation and a corresponding UI screenshot and view hierarchy; we now detail the new annotations and data post-processing. Appendix~\ref{sec:dataexappendix} contains examples of each resulting caption set, additional processing details, and a discussion on potential dataset noise.

\begin{table}[t!]
    \centering
    \setlength{\tabcolsep}{3.3pt}
    \begin{tabular}{l|c|c}
    \hline
        \multirow{2}{*}{\shortstack[l]{\textbf{Pretraining}\\ \textbf{Captioning Objective}}} & \multicolumn{2}{c}{\textbf{Pretraining Data}} \\
        \cline{2-3}
        & \# Images & \# Samples \\
        \hline
        Element (Spotlight) & 82.5M & 2.65B \\
        Element (ours) & 5,578,978 & 23,578,155\\ 
        \hline
        Element List (ours) & \multicolumn{2}{c}{5,578,978}
        \\
        Screen (ours) & \multicolumn{2}{c}{5,727,906}\\
        Textual Foresight (ours) & \multicolumn{2}{c}{2,900,572} \\
        \hline
    \end{tabular}
    \caption{Pretraining Data. We report the number of images and samples for four datasets: element captioning, as used by Spotlight~\cite{spotlight},
    reproducing element captioning data with OpenApp, a screen captioning dataset built on OpenApp, and, lastly, the Textual Foresight dataset which contains the subset of screen captions which can serve as valid foresight captions. Numbers reported for Spotlight are approximate as~\citet{spotlight} reported values with shortened notation.}
    \label{tab:datacompare}
\end{table}
\subsection{Element-Level Captions}
\label{subsec:spotlightdata}
Element captioning requires UI images with element bounding boxes and associated element captions. To obtain such pretraining samples, we process the raw OpenApp view hierarchy data to obtain every element's associated text and bounding box per image. We follow the preprocessing as detailed by~\citet{spotlight}, as we hope these annotations will approximate their work, albeit in a much smaller data regime (see Table~\ref{tab:datacompare} for sample count comparison). Element captions are obtained from all text, content description, or resource ID elements from the app view hierarchy which meet the following criteria:
\begin{enumerate}[nosep,leftmargin=*]
    \item Contains text more than one character in length, is not a URL, consists of only alphabetical characters and does not only consist of ``generic'' words (see Appendix~\ref{sec:processappendix}), and occurs at least 5 times within the respective originating dataset.
    \item Is visible, has a valid bounding box within image boundaries, and does not consist of a single pixel color (\ie, is not a color block).
\end{enumerate}
Note that we do not use an OCR model to obtain additional annotations like Spotlight did, but the AITW dataset annotations were obtained via OCR (no view hierarchy is provided for AITW). We deduplicate the resulting (app, element caption, bbox) triplets to obtain a set of unique samples.

We also include element \textit{list} captions, which operate the same way as screen captioning, but instead of having human-like natural language captions, a screen caption consists of a list of the element descriptions. 
For this formulation, we concatenate the processed element captions per screen image.
\subsection{Screen-Level Captions}
Screen captioning and Textual Foresight require (image, caption) pairs, where the caption describes the entire screen. However, to date there has been no large scale image captioning dataset for the UI domain (Screen2Words proposed by~\citet{screen2words} is used as a downstream task dataset).
To address this, we curate new OpenApp annotations with Large Language Models (LLMs).
We obtain captions for all screens by utilizing the element text available from the raw app view hierarchies. Specifically, we query GPT-3.5 Turbo~\cite{chatgpt} to obtain summaries over the elements with the following prompt:
\begin{quote}
    If an \texttt{[app package name]} app screen consisted of the following elements: \\
    \texttt{$e_0$} | \texttt{$e_1$} | ... | \texttt{$e_k$}, how would you summarize the screen? Provide a single sentence description that focuses on the functionality and category of the app given these elements. Do not repeat the app name and do not include too many specifics.
\end{quote}

\noindent and input text elements $e_k$ from each screen. In total, annotation with GPT-3.5 cost \$1,184.66 USD. These captions are then finally used as either screen captioning samples (static ($s_t$, $c_{s_t}$) pairs) or as Textual Foresight examples (interactive ($s_t$, $a_t$, $c_{s_{t+1}}$) triplets). The latter are obtained by processing valid ($s_t$, $a_t$, $s_{t+1}$) triplets from the interactive data in OpenApp. The number of images and samples for each resulting dataset is reported in Table~\ref{tab:datacompare}.

Note that the number of samples available for screen captioning is ultimately fewer than element list captioning due to different data processing (details in Appendix~\ref{sec:processappendix}). The number of samples available for Textual Foresight is almost 2x less, which is the result of numerous factors: first, we only use screens with tap actions performed and require $s_t \neq s_{t+1}$ with respect to image ID or text elements to ensure the current and next state are distinct. Second, we cannot use the final state in an action sequence as there is no following state to provide a foresight caption. Lastly, we remove samples for which we were unable to map a user interaction to a bounding box in the screen, which has been an issue in prior work as well~\cite{denoise}.
\section{Experimental Setup}
\label{sec:exps}
We now describe the new baselines made possible with the OpenApp dataset and pretraining and finetuning experimental settings.

\subsection{Baselines}
\label{subsec:static}
OpenApp contains several element and screen level caption sets that can be used to define different pretraining objectives. In addition to training Textual Foresight, we include two open-source baselines to compare to given the OpenApp data: element list captioning and screen captioning. While the OpenApp dataset includes annotations for element captioning (aiming to reproduce Spotlight with public data), it caused optimization issues with the BLIP-2 framework, possibly due to the short length of the target element captions or catastrophic forgetting. We instead compare directly to the prior published results, but still open-source these annotations for others to use, as it took substantial time to generate.

We define target captions $c_{s_t}$ for each pretraining objective (element list captioning, screen captioning, and textual foresight) below given the UI screens in OpenApp.
\[
    c_{s_t}= 
\begin{dcases}
    CAT(e_{s_t}) & \text{for      } L_{elem\_list}\\
    GPT(e_{s_t}) & \text{for      } L_{screen}, L_{foresight}
\end{dcases}
\]
\noindent As previously described, target captions $c_{s_{t+1}}$ for future screens are used to train Textual Foresight.  A benefit of our approach is that we can re-use the data from screen captioning in a new formulation, and do not require additional annotations.

Screen and element list captioning objectives can both be defined as a ``static'' loss over the current screen $s_t$:
\begin{align*}
       \hat{c}_{s_{t}} &= VLM(s_t) \\
       L_{static} &= L_{xe}(c_{s_t}, \hat{c}_{s_t})
\end{align*}
Note that we do not input a question $Q$ to our VLM when pretraining global objectives like screen and element list captioning.

\subsection{Pretraining Settings}
We use the same parameters as BLIP-2 and do not parameter tune the upstream models. Models are trained with a batch size of 100 for five epochs. The stage 2 BLIP-2 pipeline can use various LLMs; we ablated using OPT2.7, OPT6.7~\citep{zhang2022opt} and FlanT5XL~\citep{chung2022scaling}, and found early on that FlanT5 was the best language model. All results reported are with FlanT5 but additional ablations with OPT can be found in Appendix~\ref{sec:optablate}. 

Images are input to ViT at a 224x224 resolution, which is much smaller than prior work Spotlight, which input 740x740 images. High image resolutions have typically been used in prior task-specific models as well, but are hard to utilize due to current model size and memory constraints with GPUs. 

\begin{figure*}[t]
    \centering
    \includegraphics[scale=0.1635]{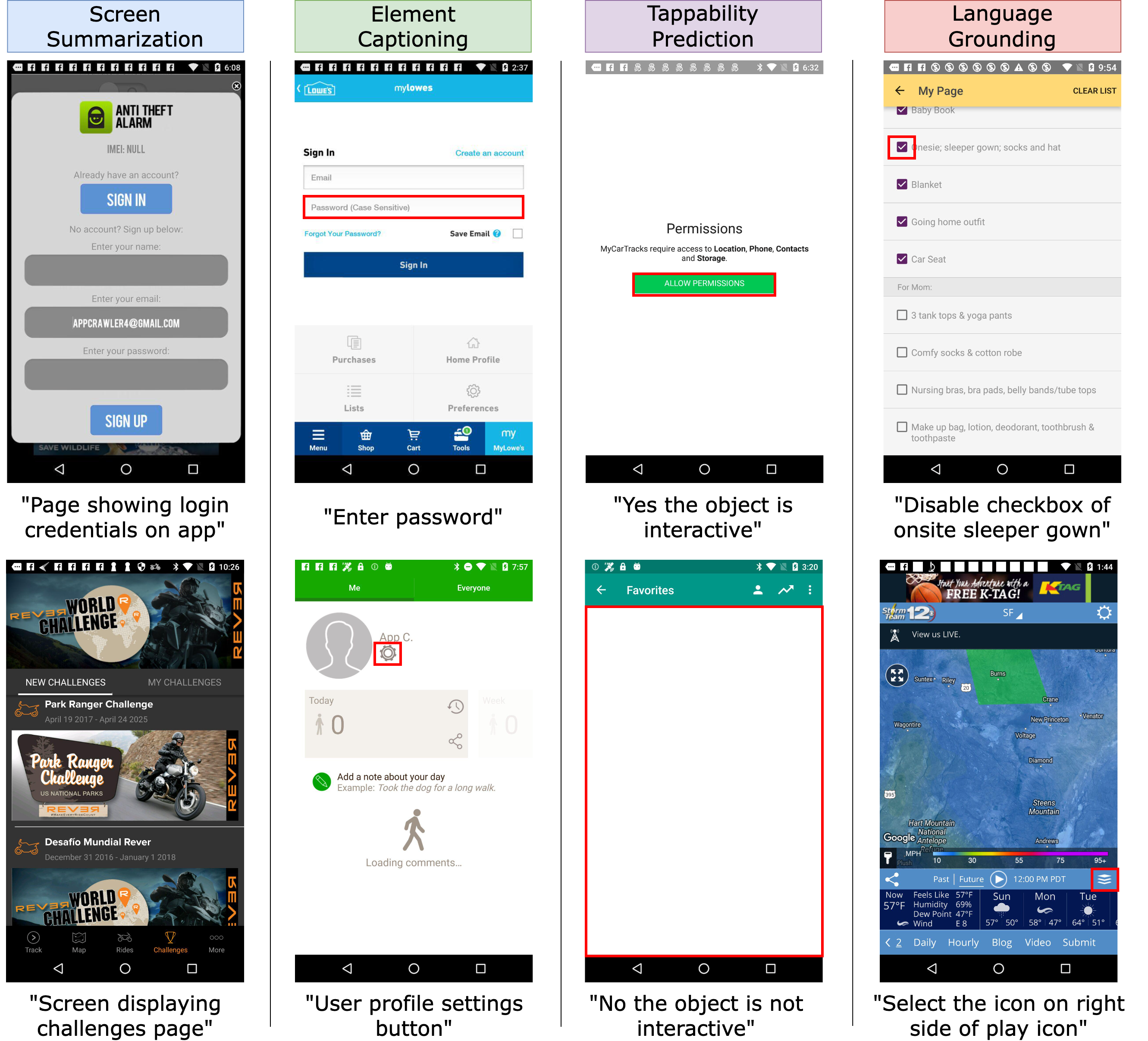}
    \caption{UI Downstream Task Examples. We illustrate samples from the Screen2Words screen summarization benchmark~\cite{screen2words}, the Widget Caption element captioning task~\cite{widgetcap}, the Tappability classification task~\cite{tappability}, and, lastly, the MUG language grounding benchmark~\cite{li2022mug}.}
    \label{fig:uitasks}
\end{figure*}

\subsection{Finetuning Settings}
Downstream models are finetuned for five epochs with a batch size of 16, and we hyperparameter tune the learning rate and number of warmup steps. We found the original learning rate 1e-5 from BLIP-2 to be most effective for the two downstream tasks with larger downstream datasets (screen summarization~\cite{screen2words} and element captioning~\cite{widgetcap}) and 5e-5 to be the most effective for the smaller tappability prediction~\cite{tappability} and language grounding datasets~\cite{li2022mug}. We selected the learning rate and number of warm up steps per downstream task via performance on the validation set (see Appendix~\ref{sec:ablateappendix} for more results). We use early stopping and report downstream results from a single run. Slightly larger image resolutions can fit into memory during finetuning, so following BLIP-2 we use the larger resolution of 364x364.

\subsection{Downstream UI Tasks and Metrics}
Our benchmark suite consists of four task datasets: screen summarization~\citep{screen2words}, element captioning~\citep{widgetcap}, tappability prediction~\citep{tappability}, and language grounding~\citep{li2022mug}. The goal of screen summarization is to provide a high level description of the entire UI screen and element captioning aims to generate captions for individual elements. Tappability prediction is the task of classifying if an element is perceived to be interactive/tappable. Lastly, the task of language grounding is to ground a single step language instruction to a UI element.
In Figure~\ref{fig:uitasks}, we illustrate samples from each downstream UI task dataset.

The primary difference from the downstream tasks used by Spotlight~\citep{spotlight} is the language grounding dataset, which was not open-sourced. We instead use the Multi-turn UI Grounding (MUG;~\citet{li2022mug}) dataset. While this dataset was proposed for multi-turn commands, approximately 80\% is single turn, and we use the full multi-turn instruction for the remaining 20\% of samples.
We describe how we formulate tappability prediction and language grounding problems as text generation tasks in Appendix~\ref{sec:modelappendix}.

For screen captioning and element captioning, we report CIDEr~\cite{cider} to be consistent with prior work, but include the more recent metrics BERTScore~\cite{hanna-bojar-2021-fine} and BLEURT-20-D12~\cite{sellam2020bleurt,pu2021learning} in Appendix~\ref{sec:ablateappendix}. For tappability prediction and language grounding F1 score and accuracy is reported, respectively. 
\section{Results}
\label{sec:res}
We now report results for generative tasks (screen and element captioning) and prediction tasks (tappability classification and language grounding).

\subsection{Generative Tasks}
\begin{table}[t!]
\setlength{\tabcolsep}{4pt}
\centering
\begin{tabular}{l|c|c|c}
\hline
\multirow{3}{*}{\bf Model} &\multicolumn{3}{c}{\bf Task} \\ 
\cline{2-4}
& Screen & Element & \multirow{2}{*}{Avg} \\
& Summ. & Caption. & \\
\hline
Screen2Words & 61.3 & -- & -- \\
Widget Caption & -- & 97.0 & -- \\
VUT & 65.6 & 99.3 & 82.5 \\
Spotlight & 106.7 & \textbf{141.8} & 124.3 \\ 
\hline
BLIP-2 (Original) & 125.1 & 121.4 & 123.3\\
\hspace{1mm} Screen Caption & 125.7 & 118.9 & 121.2 \\
\hspace{1mm} Element List & \textbf{127.9} & 121.6 & 124.8 \\
\hspace{1mm} Textual Foresight & 125.4 & \underline{128.0} & \textbf{126.7} \\
 \hline
\end{tabular}
\caption{Finetuning Generative Task Results. Prior work includes task specific methods Screen2Words~\citep{screen2words} and Widget Caption~\citep{widgetcap}, multitask model VUT~\citep{vut}, and representation learning approach Spotlight~\citep{spotlight}. All of our baselines and Textual Foresight are built upon BLIP-2~\citep{li2023blip2}. CIDEr is reported; BERTScore and BLEURT-D12 are in the Appendix.}
\label{tab:fortunexpsgen}
\end{table}
In Table~\ref{tab:fortunexpsgen}, we see the power of pretrained VLMs: BLIP-2 outperforms Spotlight with a large performance improvement on screen summarization without any further app-specific pretraining (125.1 vs.\ 106.7 CIDEr points). However, it performs worse on element captioning. This is expected, given element captioning is more domain specific and requires local understanding of the UI screen. As a result, BLIP-2 without any further pretraining trails behind Spotlight slightly on average (123.3 vs.\ 124.3). This already illustrates a trade-off, as Spotlight, which was pretrained with element captioning, intuitively does much better on this local task when evaluated downstream, while BLIP-2, which was pretrained with image captioning, does better downstream on global screen summarization.

Next, we evaluate screen captioning pretraining, made possible with our new data from OpenApp. Performance only slightly improves on screen summarization compared to BLIP-2 directly, which is surprising given the pretraining and downstream task is nearly the same. This may be, in part, due to the pretraining data: of the 5.7M unique OpenApp images, we only obtain 3.4M unique captions with GPT. \Ie, there were only 3.4M unique (app, element list) pairs, and we did not collect captions for duplicate queries.
This may result in different screens being condensed too closely in embedding space, due to incomplete text information which does not capture the ways the screens actually differ. 

In the future, querying GPT multiple times to have more unique captions may help increase caption diversity and improve performance. Another potential factor in the small performance differences could be the continued pretraining of BLIP-2 with a smaller caption dataset, which may require more careful optimization with methods like LoRA to avoid catastrophic forgetting~\cite{hu2021lora}. Unsurprisingly, we have more evidence that global captioning harms local task performance, as screen captioning actually worsens performance on element captioning compared to the baseline BLIP-2 (118.9 vs.\ 121.4).

Interestingly, the element list captioning objective, in which the global caption we aim to generate is simply the concatenated list of text elements, improves upon BLIP-2 for both tasks, and actually is the most performant on screen summarization across all pretraining objectives (bolded in the penultimate row of Table, 127.9). If the GPT-generated global screen captions were noisy or lost too much information, the raw element information may be more useful to the model. Moreover, this result demonstrates that local element information is also important to global reasoning tasks over the UI. It is surprising that list like captions proved better than natural language style sentences, suggesting quality of information retained is more crucial than style of information. The element list captioning baseline is now the first to outperform Spotlight on average across the two tasks.

Now, evaluating our proposed approach of Textual Foresight, we see a significant improvement on the element captioning task compared to our other open-source baselines (+6.4 CIDEr points compared to element list captioning, the best baseline). This is notable given that our method uses 3M \textit{fewer} samples than element list captioning, the second best method. Textual Foresight also maintains screen summarization performance, an important result that shows we can effectively blend local and global information. Ideally, we want a method which maintains the large gains on screen summarization provided by the BLIP-2 framework, while further pushing element captioning performance. Screen captioning and element list captioning maintain or slightly outperform our BLIP-2 baseline on screen summarization, but barely affect or even worsen element captioning performance. On the other hand, prior SoTA Spotlight performs the best on element captioning, but significantly worse on screen summarization, again highlighting the feature granularity trade-off. 

Instead, Textual Foresight obtains SoTA screen summarization performance.  Its largest performance impact is on element captioning, which now outperforms Spotlight on average.  In addition, our approach outperforms all other baselines in the open-source setting. In terms of data efficiency, Textual Foresight uses \textbf{28x fewer} images than Spotlight, making its gains even more impressive. We hypothesize that additional improvements could be met with our approach with access to more pretraining data or greater diversity of captions.

\begin{table}[t!]
\centering
\setlength{\tabcolsep}{4pt}
\begin{tabular}{l|c|c}
\hline
\multirow{3}{*}{\bf Model} &\multicolumn{2}{c}{\bf Task} \\ 
\cline{2-3}
& Tappability & Grounding\\
& (F1 Score) & (Accuracy) \\
\hline
Taperception & 85.5 & -- \\
Swearngin \& Li & 87.9 & -- \\
MUG & -- & \textbf{58.6} \\
VUT & 88.3 & --  \\
Spotlight & \textbf{88.4} & -- \\ 
\hline
BLIP-2 (Original) & 63.9 & 29.8 \\
\hspace{1mm} Screen Caption & 68.5 & 34.3 \\
\hspace{1mm} Element List & 67.1 & 38.2 \\
\hspace{1mm} Textual Foresight & \underline{74.2} & \underline{39.5} \\
 \hline
\end{tabular}
\caption{Finetuning Predictive Task Results. Prior work includes task specific methods Taperception~\citep{tappability},~\citet{swearngintap}, and MUG~\citep{li2022mug}, multitask model VUT~\citep{vut}, and representation learning approach Spotlight~\citep{spotlight}. All of our baselines and Textual Foresight are built upon BLIP-2~\citep{li2023blip2}.}
\label{tab:fortunexpspred}
\end{table}
\subsection{Predictive Tasks}
Now looking at classification or predictive style tasks, we report results for tappability prediction and language grounding. Textual Foresight continues to be the best open-sourced representation learning method, with improvements of up to 10.3 F1 Score and 9.7 accuracy points for tappability and grounding, respectively. Similar to our results in Table~\ref{tab:fortunexpsgen}, Textual Foresight is better than other BLIP-2 variants trained with screen and element list captioning, despite using almost half the data.

While Textual Foresight is the best in our open-source setting, these variants are ultimately less performant than prior approaches. These tasks are more challenging, as they differ more greatly from the original BLIP-2 setting of visual question answering and image captioning with natural images. Signaling the difficulty of tappability prediction and language grounding, we find all of our baseline objectives improve upon the BLIP-2 baseline model which finetunes directly on the downstream tasks. This differs from the generation-style tasks, where screen captioning actually harmed performance compared to the BLIP-2 baseline. A final consideration is the finetuning dataset size, as tappability contains 14k train samples and language grounding contains 65k, which is significantly less than the element and screen captioning datasets (138k and 78k train samples, respectively).
\section{Conclusion}
In this work we have proposed using UI actions as the bridge between local element semantics and global screen context. Specifically, we introduced a new pretraining objective, Textual Foresight, which trains a model to describe a future screen image given an action taken on the current viewed state. To train our new model we contribute a new dataset, OpenApp, which contains screen and element level captions for 5.7M app images that can be used for training several baselines. We are the first to provide an open-source app dataset for UI representation learning and evaluate on a standardized downstream benchmark. Our Textual Foresight approach can use only a subset of this data and on average outperforms not only our open-source benchmarks, but also prior state-of-the-art method Spotlight on generation tasks, while using 2x less data than open-source baselines, and 28x less data than prior state-of-the-art.

\section{Limitations}
In this work we curate new data for the proposed OpenApp dataset in part with LLMs like GPT3.5 Turbo. As a result, our image captions do not necessarily capture the full image content accurately, or may lose information that would otherwise be helpful for representation learning. While other works have utilized pseudo summaries or automatic summarizations~\cite{Narasimhan2022TLDWSI,burns-etal-2023-suite}, it is important to note that human annotation or verification of our dataset could improve its quality in future work.

Additionally, as discussed in our results, all of our baselines and Textual Foresight fall short for prediction style tasks. Given how low BLIP-2 (Original) baseline performance is, it is possibly a limitation of the model framework, along with other factors like the scale of our pretraining data or size of finetuning data. Currently, our work is most effective for captioning and summarization style tasks, but we hope our full benchmark will allow for fair comparison in future research and new open source tools, as prior representation learning approaches did not provide any resources for reproducing their methods. We also did not try all possible combinations of our pretraining objectives due to computational and time constraints.

Lastly, while it is possible that the mobile app UI data includes non-English content, they were designed and built as English datasets. As a result, the models trained for various tasks are only reliable for English as of now. In future work, it would be important to both intentionally curate multilingual UI data, as well as quantify how much data in existing sources in already multilingual (\eg, there may be spurious text or ads in other languages, for example).

\section{Ethics}
Curating data and automating tasks in the UI domain requires consideration of user privacy and safety, as well as user demographic. We do not collect any new mobile app action sequences, as we only build new annotations on top of existing open source datasets. As a result, we do not introduce any new ethical issues related to the data source. However, when modeling downstream tasks, there are inherent risks with models that perform tasks on behalf of humans, such as language grounding (in which a user instruction is automated on their behalf). There are many situations in which a user would not be able to double check the model output, and for this reason additional work is needed to provide explainable predictions and only automate tasks when there is high model confidence. This concern is less applicable to captioning and summarization UI problems. 

With respect to privacy, people that use assistive technology or human-in-the-loop tools already expose P.I.I. information to be able to use mobile apps~\cite{uncomf,privacy}. Still, an ethical concern that persists is to ensure the models we train do not retain any user-specific information if they are finetuned or personalized for individuals. This is out of scope for our work, but we note that the UI data within OpenApp was created with anonymous login credentials when originally annotated.

\section*{Acknowledgements}

This work is supported, in part, by the Google Ph.D. Fellowship program.
\bibliography{main}

\begin{thebibliography}{49}
\expandafter\ifx\csname natexlab\endcsname\relax\def\natexlab#1{#1}\fi

\bibitem[{Ahmed et~al.(2015)Ahmed, Hoyle, Connelly, Crandall, and Kapadia}]{privacy}
Tousif Ahmed, Roberto Hoyle, Kay Connelly, David Crandall, and Apu Kapadia. 2015.
\newblock \href {https://doi.org/10.1145/2702123.2702334} {Privacy concerns and behaviors of people with visual impairments}.
\newblock In \emph{Proceedings of the 33rd Annual ACM Conference on Human Factors in Computing Systems}, CHI '15, page 3523–3532, New York, NY, USA. Association for Computing Machinery.

\bibitem[{Akter et~al.(2020)Akter, Dosono, Ahmed, Kapadia, and Semaan}]{uncomf}
Taslima Akter, Bryan Dosono, Tousif Ahmed, Apu Kapadia, and Bryan Semaan. 2020.
\newblock “i am uncomfortable sharing what i can't see”: privacy concerns of the visually impaired with camera based assistive applications.
\newblock In \emph{Proceedings of the 29th USENIX Conference on Security Symposium}, SEC'20, USA. USENIX Association.

\bibitem[{Babaeizadeh et~al.(2017)Babaeizadeh, Finn, Erhan, Campbell, and Levine}]{stochasticvar}
Mohammad Babaeizadeh, Chelsea Finn, Dumitru Erhan, Roy~H. Campbell, and Sergey Levine. 2017.
\newblock \href {http://arxiv.org/abs/1710.11252} {Stochastic variational video prediction}.
\newblock \emph{CoRR}, abs/1710.11252.

\bibitem[{Bai et~al.(2021)Bai, Zang, Xu, Sunkara, Rastogi, Chen, and y~Arcas}]{bai2021uibert}
Chongyang Bai, Xiaoxue Zang, Ying Xu, Srinivas Sunkara, Abhinav Rastogi, Jindong Chen, and Blaise~Aguera y~Arcas. 2021.
\newblock \href {http://arxiv.org/abs/2107.13731} {Uibert: Learning generic multimodal representations for ui understanding}.

\bibitem[{Burns et~al.(2022)Burns, Arsan, Agrawal, Kumar, Saenko, and Plummer}]{burns2022motifvln}
Andrea Burns, Deniz Arsan, Sanjna Agrawal, Ranjitha Kumar, Kate Saenko, and Bryan~A. Plummer. 2022.
\newblock A dataset for interactive vision language navigation with unknown command feasibility.
\newblock In \emph{European Conference on Computer Vision (ECCV)}.

\bibitem[{Burns et~al.(2023)Burns, Srinivasan, Ainslie, Brown, Plummer, Saenko, Ni, and Guo}]{burns-etal-2023-suite}
Andrea Burns, Krishna Srinivasan, Joshua Ainslie, Geoff Brown, Bryan Plummer, Kate Saenko, Jianmo Ni, and Mandy Guo. 2023.
\newblock \href {https://doi.org/10.18653/v1/2023.emnlp-main.119} {A suite of generative tasks for multi-level multimodal webpage understanding}.
\newblock In \emph{Proceedings of the 2023 Conference on Empirical Methods in Natural Language Processing}, pages 1917--1947, Singapore. Association for Computational Linguistics.

\bibitem[{Chung et~al.(2022)Chung, Hou, Longpre, Zoph, Tay, Fedus, Li, Wang, Dehghani, Brahma, Webson, Gu, Dai, Suzgun, Chen, Chowdhery, Castro-Ros, Pellat, Robinson, Valter, Narang, Mishra, Yu, Zhao, Huang, Dai, Yu, Petrov, Chi, Dean, Devlin, Roberts, Zhou, Le, and Wei}]{chung2022scaling}
Hyung~Won Chung, Le~Hou, Shayne Longpre, Barret Zoph, Yi~Tay, William Fedus, Yunxuan Li, Xuezhi Wang, Mostafa Dehghani, Siddhartha Brahma, Albert Webson, Shixiang~Shane Gu, Zhuyun Dai, Mirac Suzgun, Xinyun Chen, Aakanksha Chowdhery, Alex Castro-Ros, Marie Pellat, Kevin Robinson, Dasha Valter, Sharan Narang, Gaurav Mishra, Adams Yu, Vincent Zhao, Yanping Huang, Andrew Dai, Hongkun Yu, Slav Petrov, Ed~H. Chi, Jeff Dean, Jacob Devlin, Adam Roberts, Denny Zhou, Quoc~V. Le, and Jason Wei. 2022.
\newblock \href {http://arxiv.org/abs/2210.11416} {Scaling instruction-finetuned language models}.

\bibitem[{Dai et~al.(2023)Dai, Li, Li, Tiong, Zhao, Wang, Li, Fung, and Hoi}]{dai2023instructblip}
Wenliang Dai, Junnan Li, Dongxu Li, Anthony Meng~Huat Tiong, Junqi Zhao, Weisheng Wang, Boyang Li, Pascale Fung, and Steven Hoi. 2023.
\newblock \href {http://arxiv.org/abs/2305.06500} {Instructblip: Towards general-purpose vision-language models with instruction tuning}.

\bibitem[{Desai and Johnson(2021)}]{desai2021virtex}
Karan Desai and Justin Johnson. 2021.
\newblock {VirTex: Learning Visual Representations from Textual Annotations}.
\newblock In \emph{CVPR}.

\bibitem[{Dogruer et~al.(2011)Dogruer, Eyyam, and Menevis}]{DOGRUER2011606}
Nazan Dogruer, Ramadan Eyyam, and Ipek Menevis. 2011.
\newblock {The Use of the Internet for Educational Purposes}.
\newblock In \emph{Procedia - Social and Behavioral Sciences}.

\bibitem[{Dosovitskiy et~al.(2021)Dosovitskiy, Beyer, Kolesnikov, Weissenborn, Zhai, Unterthiner, Dehghani, Minderer, Heigold, Gelly, Uszkoreit, and Houlsby}]{dosovitskiy2020vit}
Alexey Dosovitskiy, Lucas Beyer, Alexander Kolesnikov, Dirk Weissenborn, Xiaohua Zhai, Thomas Unterthiner, Mostafa Dehghani, Matthias Minderer, Georg Heigold, Sylvain Gelly, Jakob Uszkoreit, and Neil Houlsby. 2021.
\newblock An image is worth 16x16 words: Transformers for image recognition at scale.
\newblock \emph{ICLR}.

\bibitem[{Finn and Levine(2017)}]{foresight}
Chelsea Finn and Sergey Levine. 2017.
\newblock \href {https://doi.org/10.1109/ICRA.2017.7989324} {Deep visual foresight for planning robot motion}.
\newblock In \emph{International Conference on Robotics and Automation (ICRA)}, pages 2786--2793.

\bibitem[{Fok et~al.(2022)Fok, Zhong, Ross, Fogarty, and Wobbrock}]{fok_longa11y_2022}
Raymond Fok, Mingyuan Zhong, Anne~Spencer Ross, James Fogarty, and Jacob~O. Wobbrock. 2022.
\newblock \href {https://doi.org/10.1145/3491102.3502143} {A large-scale longitudinal analysis of missing label accessibility failures in android apps}.
\newblock In \emph{ACM Conference on Human Factors in Computing Systems}, CHI '22, New York, NY, USA. Association for Computing Machinery.

\bibitem[{Hanna and Bojar(2021)}]{hanna-bojar-2021-fine}
Michael Hanna and Ond{\v{r}}ej Bojar. 2021.
\newblock \href {https://aclanthology.org/2021.wmt-1.59} {A fine-grained analysis of {BERTS}core}.
\newblock In \emph{Proceedings of the Sixth Conference on Machine Translation}, pages 507--517, Online. Association for Computational Linguistics.

\bibitem[{He et~al.(2021)He, Sunkara, Zang, Xu, Liu, Wichers, Schubiner, Lee, and Chen}]{actionbert}
Zecheng He, Srinivas Sunkara, Xiaoxue Zang, Ying Xu, Lijuan Liu, Nevan Wichers, Gabriel Schubiner, Ruby Lee, and Jindong Chen. 2021.
\newblock \href {https://doi.org/10.1609/aaai.v35i7.16741} {Actionbert: Leveraging user actions for semantic understanding of user interfaces}.
\newblock \emph{Proceedings of the AAAI Conference on Artificial Intelligence}, 35(7):5931--5938.

\bibitem[{Hoque et~al.(2020)Hoque, Seita, Balakrishna, Ganapathi, Tanwani, Jamali, Yamane, Iba, and Goldberg}]{robotforesight}
Ryan Hoque, Daniel Seita, Ashwin Balakrishna, Aditya Ganapathi, Ajay~Kumar Tanwani, Nawid Jamali, Katsu Yamane, Soshi Iba, and Ken Goldberg. 2020.
\newblock \href {https://doi.org/10.48550/ARXIV.2003.09044} {Visuospatial foresight for multi-step, multi-task fabric manipulation}.

\bibitem[{Hu et~al.(2021)Hu, Shen, Wallis, Allen-Zhu, Li, Wang, Wang, and Chen}]{hu2021lora}
Edward~J. Hu, Yelong Shen, Phillip Wallis, Zeyuan Allen-Zhu, Yuanzhi Li, Shean Wang, Lu~Wang, and Weizhu Chen. 2021.
\newblock \href {http://arxiv.org/abs/2106.09685} {Lora: Low-rank adaptation of large language models}.

\bibitem[{Koh et~al.(2021)Koh, Lee, Yang, Baldridge, and Anderson}]{koh2021pathdreamer}
Jing~Yu Koh, Honglak Lee, Yinfei Yang, Jason Baldridge, and Peter Anderson. 2021.
\newblock Pathdreamer: A world model for indoor navigation.
\newblock In \emph{ICCV}.

\bibitem[{Koh et~al.(2024)Koh, Lo, Jang, Duvvur, Lim, Huang, Neubig, Zhou, Salakhutdinov, and Fried}]{koh2024visualwebarena}
Jing~Yu Koh, Robert Lo, Lawrence Jang, Vikram Duvvur, Ming~Chong Lim, Po-Yu Huang, Graham Neubig, Shuyan Zhou, Ruslan Salakhutdinov, and Daniel Fried. 2024.
\newblock Visualwebarena: Evaluating multimodal agents on realistic visual web tasks.
\newblock \emph{arXiv preprint arXiv:2401.13649}.

\bibitem[{Lee et~al.(2018)Lee, Zhang, Ebert, Abbeel, Finn, and Levine}]{stochasticadv}
Alex~X. Lee, Richard Zhang, Frederik Ebert, Pieter Abbeel, Chelsea Finn, and Sergey Levine. 2018.
\newblock \href {https://doi.org/10.48550/ARXIV.1804.01523} {Stochastic adversarial video prediction}.

\bibitem[{Li et~al.(2022{\natexlab{a}})Li, Baechler, Tragut, and Li}]{denoise}
Gang Li, Gilles Baechler, Manuel Tragut, and Yang Li. 2022{\natexlab{a}}.
\newblock \href {https://doi.org/10.1145/3491102.3502042} {Learning to denoise raw mobile ui layouts for improving datasets at scale}.
\newblock In \emph{Proceedings of the 2022 CHI Conference on Human Factors in Computing Systems}, CHI '22, New York, NY, USA. Association for Computing Machinery.

\bibitem[{Li and Li(2023)}]{spotlight}
Gang Li and Yang Li. 2023.
\newblock Spotlight: Mobile ui understanding using vision-language models with a focus.
\newblock In \emph{International Conference on Learning Representations (ICLR)}.

\bibitem[{Li et~al.(2023)Li, Li, Savarese, and Hoi}]{li2023blip2}
Junnan Li, Dongxu Li, Silvio Savarese, and Steven Hoi. 2023.
\newblock \href {http://arxiv.org/abs/2301.12597} {Blip-2: Bootstrapping language-image pre-training with frozen image encoders and large language models}.

\bibitem[{Li et~al.(2022{\natexlab{b}})Li, Li, Zheng, Wang, and Li}]{li2022mug}
Tao Li, Gang Li, Jingjie Zheng, Purple Wang, and Yang Li. 2022{\natexlab{b}}.
\newblock \href {http://arxiv.org/abs/2209.15099} {Mug: Interactive multimodal grounding on user interfaces}.

\bibitem[{Li et~al.(2021{\natexlab{a}})Li, Popowski, Mitchell, and Myers}]{screen2vec}
Toby Jia-Jun Li, Lindsay Popowski, Tom~M. Mitchell, and Brad~A. Myers. 2021{\natexlab{a}}.
\newblock Screen2vec: Semantic embedding of gui screens and gui components.
\newblock In \emph{Proceedings of the SIGCHI Conference on Human Factors in Computing Systems}, CHI '21.

\bibitem[{Li et~al.(2020)Li, Li, He, Zheng, Li, and Guan}]{widgetcap}
Yang Li, Gang Li, Luheng He, Jingjie Zheng, Hong Li, and Zhiwei Guan. 2020.
\newblock \href {https://doi.org/10.48550/ARXIV.2010.04295} {Widget captioning: Generating natural language description for mobile user interface elements}.

\bibitem[{Li et~al.(2021{\natexlab{b}})Li, Li, Zhou, Dehghani, and Gritsenko}]{vut}
Yang Li, Gang Li, Xin Zhou, Mostafa Dehghani, and Alexey~A. Gritsenko. 2021{\natexlab{b}}.
\newblock \href {http://arxiv.org/abs/2112.05692} {{VUT:} versatile {UI} transformer for multi-modal multi-task user interface modeling}.
\newblock \emph{CoRR}, abs/2112.05692.

\bibitem[{Nair et~al.(2018)Nair, Pong, Dalal, Bahl, Lin, and Levine}]{NEURIPS2018_7ec69dd4}
Ashvin~V Nair, Vitchyr Pong, Murtaza Dalal, Shikhar Bahl, Steven Lin, and Sergey Levine. 2018.
\newblock \href {https://proceedings.neurips.cc/paper/2018/file/7ec69dd44416c46745f6edd947b470cd-Paper.pdf} {Visual reinforcement learning with imagined goals}.
\newblock In \emph{Advances in Neural Information Processing Systems}, volume~31. Curran Associates, Inc.

\bibitem[{Nair and Finn(2019)}]{hierforesight}
Suraj Nair and Chelsea Finn. 2019.
\newblock \href {https://doi.org/10.48550/ARXIV.1909.05829} {Hierarchical foresight: Self-supervised learning of long-horizon tasks via visual subgoal generation}.

\bibitem[{Nair et~al.(2020)Nair, Savarese, and Finn}]{pmlr-v119-nair20a}
Suraj Nair, Silvio Savarese, and Chelsea Finn. 2020.
\newblock \href {https://proceedings.mlr.press/v119/nair20a.html} {Goal-aware prediction: Learning to model what matters}.
\newblock In \emph{Proceedings of the 37th International Conference on Machine Learning}, volume 119 of \emph{Proceedings of Machine Learning Research}, pages 7207--7219. PMLR.

\bibitem[{Narasimhan et~al.(2022)Narasimhan, Nagrani, Sun, Rubinstein, Darrell, Rohrbach, and Schmid}]{Narasimhan2022TLDWSI}
Medhini~G. Narasimhan, Arsha Nagrani, Chen Sun, Michael Rubinstein, Trevor Darrell, Anna Rohrbach, and Cordelia Schmid. 2022.
\newblock \href {https://api.semanticscholar.org/CorpusID:251564419} {Tl;dw? summarizing instructional videos with task relevance \& cross-modal saliency}.
\newblock In \emph{European Conference on Computer Vision}.

\bibitem[{OpenAI(2022)}]{chatgpt}
OpenAI. 2022.
\newblock \href {https://openai.com/blog/chatgpt} {{ChatGPT}}.

\bibitem[{Pu et~al.(2021)Pu, Chung, Parikh, Gehrmann, and Sellam}]{pu2021learning}
Amy Pu, Hyung~Won Chung, Ankur~P Parikh, Sebastian Gehrmann, and Thibault Sellam. 2021.
\newblock Learning compact metrics for mt.
\newblock In \emph{Proceedings of EMNLP}.

\bibitem[{Raffel et~al.(2020)Raffel, Shazeer, Roberts, Lee, Narang, Matena, Zhou, Li, and Liu}]{2020t5}
Colin Raffel, Noam Shazeer, Adam Roberts, Katherine Lee, Sharan Narang, Michael Matena, Yanqi Zhou, Wei Li, and Peter~J. Liu. 2020.
\newblock \href {http://jmlr.org/papers/v21/20-074.html} {Exploring the limits of transfer learning with a unified text-to-text transformer}.
\newblock \emph{Journal of Machine Learning Research}, 21(140):1--67.

\bibitem[{Rawles et~al.(2023)Rawles, Li, Rodriguez, Riva, and Lillicrap}]{rawles2023android}
Christopher Rawles, Alice Li, Daniel Rodriguez, Oriana Riva, and Timothy Lillicrap. 2023.
\newblock \href {http://arxiv.org/abs/2307.10088} {Android in the wild: A large-scale dataset for android device control}.

\bibitem[{Schoop et~al.(2022)Schoop, Zhou, Li, Chen, Hartmann, and Li}]{tappability}
Eldon Schoop, Xin Zhou, Gang Li, Zhourong Chen, Bjoern Hartmann, and Yang Li. 2022.
\newblock \href {https://doi.org/10.1145/3491102.3517497} {Predicting and explaining mobile ui tappability with vision modeling and saliency analysis}.
\newblock In \emph{Proceedings of the 2022 CHI Conference on Human Factors in Computing Systems}, CHI '22, New York, NY, USA. Association for Computing Machinery.

\bibitem[{Sellam et~al.(2020)Sellam, Das, and Parikh}]{sellam2020bleurt}
Thibault Sellam, Dipanjan Das, and Ankur~P Parikh. 2020.
\newblock Bleurt: Learning robust metrics for text generation.
\newblock In \emph{Proceedings of ACL}.

\bibitem[{Srinivasan et~al.(2021)Srinivasan, Raman, Chen, Bendersky, and Najork}]{wit}
Krishna Srinivasan, Karthik Raman, Jiecao Chen, Michael Bendersky, and Marc Najork. 2021.
\newblock \href {https://doi.org/10.1145/3404835.3463257} {Wit: Wikipedia-based image text dataset for multimodal multilingual machine learning}.
\newblock In \emph{Proceedings of the 44th International ACM SIGIR Conference on Research and Development in Information Retrieval}, SIGIR '21, page 2443–2449, New York, NY, USA. Association for Computing Machinery.

\bibitem[{Swearngin and Li(2019)}]{swearngintap}
Amanda Swearngin and Yang Li. 2019.
\newblock \href {https://doi.org/10.1145/3290605.3300305} {Modeling mobile interface tappability using crowdsourcing and deep learning}.
\newblock In \emph{Proceedings of the 2019 CHI Conference on Human Factors in Computing Systems}, CHI '19, page 1–11, New York, NY, USA. Association for Computing Machinery.

\bibitem[{Vaswani et~al.(2017)Vaswani, Shazeer, Parmar, Uszkoreit, Jones, Gomez, Kaiser, and Polosukhin}]{transformer}
Ashish Vaswani, Noam Shazeer, Niki Parmar, Jakob Uszkoreit, Llion Jones, Aidan~N Gomez, Lukasz Kaiser, and Illia Polosukhin. 2017.
\newblock Attention is all you need.
\newblock In \emph{Advances in Neural Information Processing Systems}, pages 6000--6010.

\bibitem[{Vedantam et~al.(2015)Vedantam, Zitnick, and Parikh}]{cider}
Ramakrishna Vedantam, C.~Lawrence Zitnick, and Devi Parikh. 2015.
\newblock {CIDEr: Consensus-based Image Description Evaluation}.
\newblock In \emph{IEEE/CVF Conference on Computer Vision and Pattern Recognition (CVPR)}.

\bibitem[{Vinyals et~al.(2015)Vinyals, Toshev, Bengio, and Erhan}]{vinyalsCVPR2015}
Oriol Vinyals, Alexander Toshev, Samy Bengio, and Dumitru Erhan. 2015.
\newblock Show and tell: A neural image caption generator.
\newblock In \emph{CVPR}.

\bibitem[{Vtyurina et~al.(2019)Vtyurina, Fourney, Morris, Findlater, and White}]{screenreader}
Alexandra Vtyurina, Adam Fourney, Meredith~Ringel Morris, Leah Findlater, and Ryen~W. White. 2019.
\newblock Bridging screen readers and voice assistants for enhanced eyes-free web search.
\newblock In \emph{International ACM SIGACCESS Conference on Computers and Accessibility (ASSETS)}.

\bibitem[{Wang et~al.(2021)Wang, Li, Zhou, Chen, Grossman, and Li}]{screen2words}
Bryan Wang, Gang Li, Xin Zhou, Zhourong Chen, Tovi Grossman, and Yang Li. 2021.
\newblock \href {https://doi.org/10.1145/3472749.3474765} {Screen2words: Automatic mobile ui summarization with multimodal learning}.
\newblock In \emph{The 34th Annual ACM Symposium on User Interface Software and Technology}, UIST '21, page 498–510, New York, NY, USA. Association for Computing Machinery.

\bibitem[{Wang et~al.(2018)Wang, Xiong, Wang, and Wang}]{foresightvln}
Xin Wang, Wenhan Xiong, Hongmin Wang, and William Wang. 2018.
\newblock Look before you leap: Bridging model-free and model-based reinforcement learning for planned-ahead vision-and-language navigation.
\newblock In \emph{European Conference on Computer Vision (ECCV)}.

\bibitem[{Yao et~al.(preprint)Yao, Chen, Yang, and Narasimhan}]{yao2022webshop}
Shunyu Yao, Howard Chen, John Yang, and Karthik Narasimhan. preprint.
\newblock Webshop: Towards scalable real-world web interaction with grounded language agents.
\newblock In \emph{ArXiv}.

\bibitem[{Yen-Chen et~al.(2020)Yen-Chen, Bauza, and Isola}]{pmlr-v100-yen-chen20a}
Lin Yen-Chen, Maria Bauza, and Phillip Isola. 2020.
\newblock \href {https://proceedings.mlr.press/v100/yen-chen20a.html} {Experience-embedded visual foresight}.
\newblock In \emph{Proceedings of the Conference on Robot Learning}, volume 100 of \emph{Proceedings of Machine Learning Research}, pages 1015--1024. PMLR.

\bibitem[{Zhang et~al.(2022)Zhang, Roller, Goyal, Artetxe, Chen, Chen, Dewan, Diab, Li, Lin, Mihaylov, Ott, Shleifer, Shuster, Simig, Koura, Sridhar, Wang, and Zettlemoyer}]{zhang2022opt}
Susan Zhang, Stephen Roller, Naman Goyal, Mikel Artetxe, Moya Chen, Shuohui Chen, Christopher Dewan, Mona Diab, Xian Li, Xi~Victoria Lin, Todor Mihaylov, Myle Ott, Sam Shleifer, Kurt Shuster, Daniel Simig, Punit~Singh Koura, Anjali Sridhar, Tianlu Wang, and Luke Zettlemoyer. 2022.
\newblock \href {http://arxiv.org/abs/2205.01068} {Opt: Open pre-trained transformer language models}.

\bibitem[{Zhao et~al.(2016)Zhao, Ramos, Tao, Jiang, Li, Wu, Pan, and Dey}]{appbehavior}
Sha Zhao, Julian Ramos, Jianrong Tao, Ziwen Jiang, Shijian Li, Zhaohui Wu, Gang Pan, and Anind~K. Dey. 2016.
\newblock {Discovering Different Kinds of Smartphone Users through Their Application Usage Behaviors}.
\newblock In \emph{International ACM Joint Conference on Pervasive and Ubiquitous Computing (UbiComp)}.

\end{thebibliography}

\appendix

\newpage 

\section{Data Processing Details}
\label{sec:processappendix}
We include additional details for the data processing used to obtain each OpenApp captioning sample set from the raw view hierarchy data. We release all of our code, including the data processing pipelines, so others can reproduce our work or modify our pipeline as needed.

\subsection{Element Captioning Data}
As discussed in the main text, our aim in generating the element captioning data was to reproduce a dataset as similar to Spotlight's as possible. Thus, we followed their same data processing rules. Element captions are obtained from all text, content description, or resource ID fields from the app view hierarchy elements which meet the below criteria:
\begin{enumerate}
    \item Contains text more than one character in length, is not a URL, consists of only alphabetical characters and does not only consist of ``generic'' words, and occurs at least 5 times within the respective originating dataset.
    \item Is visible, has a valid bounding box within image boundaries, and does not consist of a single pixel color (\ie, is not a color block).
\end{enumerate}
The list of generic words is: action, bar, menu, title, and, ans,
app, icon, name, arg, background, element, btn, but, bottom, button, content, desc, text, item, empty, fab, image, grid, header, img, imgfile, lbutton, label, letter, list, view, pic, placeholder, random, row, single, raw, small, large, sub, template, navbar, banner, test, textinput, error, texto, todo, toolbar, tool, track, txt, unknown, stub, web, left, right, tlb, nan, page, feature, menugrid, picture, tabs, number, node, iconimage, entity, webview, heading, logo, tbl, tab, primary, and footer per Spotlight. Lastly, all fields were made lowercase.

These stringent processing rules are needed due to potential noise and inaccuracies in the app view hierarchy. In particular, ensuring the bounding boxes lie within image boundaries is important for any localized task like element captioning or textual foresight.

\subsection{Element List Captioning Data}
Our element list captioning dataset concatenates all of the element text per screen from the element captioning dataset. The elements are joined by commas. This results in a screen captioning-style task where the captions to decode are element list strings instead of natural language captions.

\subsection{Screen Captioning Data}
Both our screen captioning and textual foresight captions are obtained in the same manner with the GPT-3.5 Turbo API. As mentioned in the main text, we generate text prompts for each screen in OpenApp to obtain a screen caption. Specifically, we input:
\begin{quote}
    If an \texttt{[app package name]} app screen consisted of the following elements: \\
    \texttt{$e_0$} | \texttt{$e_1$} | ... | \texttt{$e_k$}, how would you summarize the screen? Provide a single sentence description that focuses on the functionality and category of the app given these elements. Do not repeat the app name and do not include too many specifics.
\end{quote}
and query GPT-3.5 with the set of \textit{unique} samples. This means if multiple different screens from the same app had the same list of cleaned elements $e_k$, we only queried GPT-3.5 once for them. In the future, augmentations of the same caption could be obtained by re-querying the model again. There currently is no way to ``seed'' the GPT models, meaning that even for the exact same input and model checkpoint, the output is often different when the API is called more than once for a particular sample. Setting the temperature to zero does not fully control the model output, either.

For the screen-level caption sets, we use a slightly different set of processing steps to clean the raw view hierarchy elements $e_k$. First, we chose to not use resource ID text fields as valid elements due to them being noisy and more like generic metadata, proving less useful for reasoning about the specific UI screen. We also retain upper case text as this could be helpful to the GPT model.

\subsection{Textual Foresight Data}
The captions that are used for textual foresight come from the same GPT-3.5 outputs as described in the prior section. However, what differs is which screens we can utilize. We choose to only use screens that have tap actions performed on them, as swiping and editing text fields on the UI may not change the UI enough to warrant a foresight caption which differs significantly from the current screen's caption.

In any mobile app dataset containing action sequences, a key part of using the user action annotations is mapping the screen interactions to view hierarchy bounding boxes. The user actions and view hierarchy elements exist in different scales and must be normalized to be mapped to one another. While an action should exactly match one UI element, there are times when it matches zero. This can occur due to a human's click being located slightly outside of the true bounding box. Additionally, this occurs more often for the Android In The Wild dataset within OpenApp, due to it using OCR. Specifically, sometimes the OCR does not include strictly visual elements or has other failure cases. 

To address this, for the subset of actions that are not initially within an element's bounding box, we try to enlarge the view hierarchy bounds by small amounts until the action coordinate falls within one. If this is ineffective at a certain threshold, we will instead create a square box of 65x65 pixels centered around the user action location. This occurs for various edge cases like keyboards, calculators, icons, and the phone dialer, which correspond to no known element in the view hierarchy or detected OCR.

We also specially deal with other edge cases, \eg, if we find an action is clicking back on the UI banner, we do not include it. Additionally, there are cases when an action location is within more than one bounding box, as the bounding boxes can be overlapping at times. Of the matching bounding boxed, we will select the one with lowest euclidean distance to its midpoint with the smallest area.

All of the code used to capture these edge cases and process them is included in the \href{https://github.com/aburns4/textualforesight}{GitHub repository}.

\section{Dataset Examples}
\label{sec:dataexappendix}
\begin{figure*}
\centering
    \includegraphics[scale=0.0605]{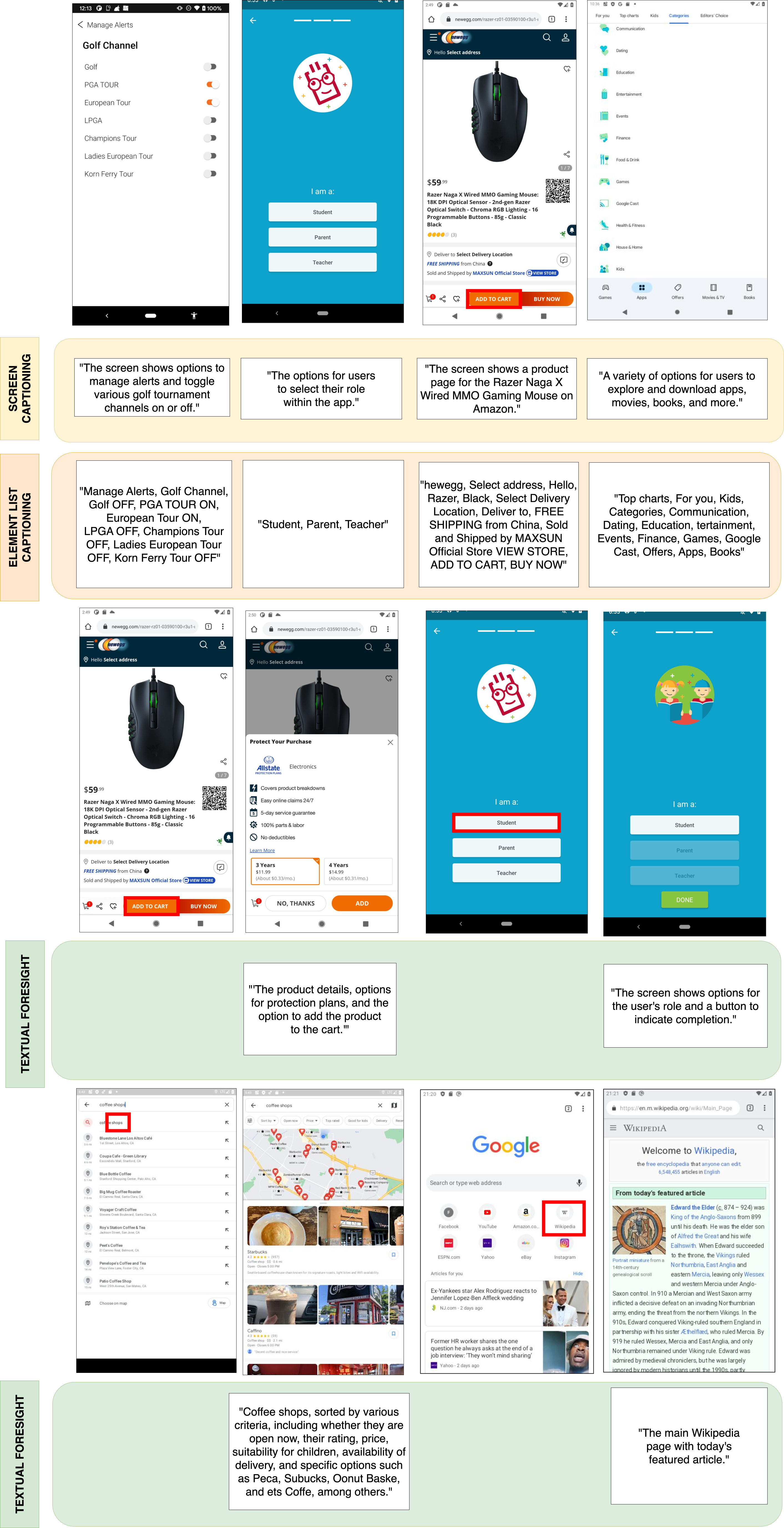}
    \caption{Examples from the OpenApp dataset. We show example new captions we build for OpenApp from the element captioning, element list captioning, screen captioning, and textual foresight sample sets.}
    \label{fig:openappex}
\end{figure*}

In Figure~\ref{fig:openappex} we include example images and captions for all caption sets in our OpenApp dataset: element captioning, element list captioning, screen captioning, and textual foresight. Element captioning would result in separate samples for every text element in the element list captions (each element is comma separated). For example, for the user choice page in blue (first row, second example of Figure~\ref{fig:openappex}), the element list caption is simply ``Student, Parent, Teacher'' and the corresponding element level captions would be ``Student,'' ``Parent,'' and ``Teacher.'' The screen captioning set are the result of our separate element processing pipeline and GPT3.5 Turbo querying.

Lastly, we illustrate four examples of textual foresight. We show both the input image and sequential image (left and right respectively) for visualization purposes; we only input the current screen and our action question to generate the foresight caption. We also highlight the action element in red for clarity (\ie, these red bounding boxes are not actually on the input images). We include foresight captions underneath the next screen in Figure~\ref{fig:openappex}. Interestingly, even when foresight captions do not extend greatly beyond the action element's semantics, they can serve as a proxy for a more descriptive element caption (see the bottom right Wikipedia example).

\section{Dataset Noise}
In our OpenApp dataset, there are two potential sources of noise. First, as partially discussed in Appendix \ref{sec:processappendix}, to have questions with local action or element grounding information (for textual foresight and element captioning objectives, respectively), human actions on the UI screen have to be matched with backend view hierarchy bounding boxes. There are a subset of cases where there is not an exact 1-1 mapping between the two, and we either find a nearby bounding box or create a new one around the action coordinate. This process is imperfect, but we manually inspected around 100 processed samples per dataset in OpenApp to ensure reasonable quality. For our textual foresight approach, a perfect localization on the screen is also not always needed.

The second potential source of dataset noise comes from using GPT-3.5 Turbo to generate captions and meaningfully aggregate view hierarchy element text. While it is unlikely for the GPT to generate something not related to the screen inputs, it is possible that the resulting summary misses the most salient screen details that should appear in an image caption. This can happen as a result of many distractor elements which obfuscate the true focus of the screen. 

While it is possible GPT-4 could better produce captions, or that GPT-3.5 would do better by inputting the entire raw view hierarchy (such that all structure and metadata is retained), this would be prohibitively expensive. The GPT-4 API is significantly more expensive than earlier models, and price is determined by both input and output text length (\ie, number of tokens).

\begin{figure*}
    \centering
    \includegraphics[scale=0.21]{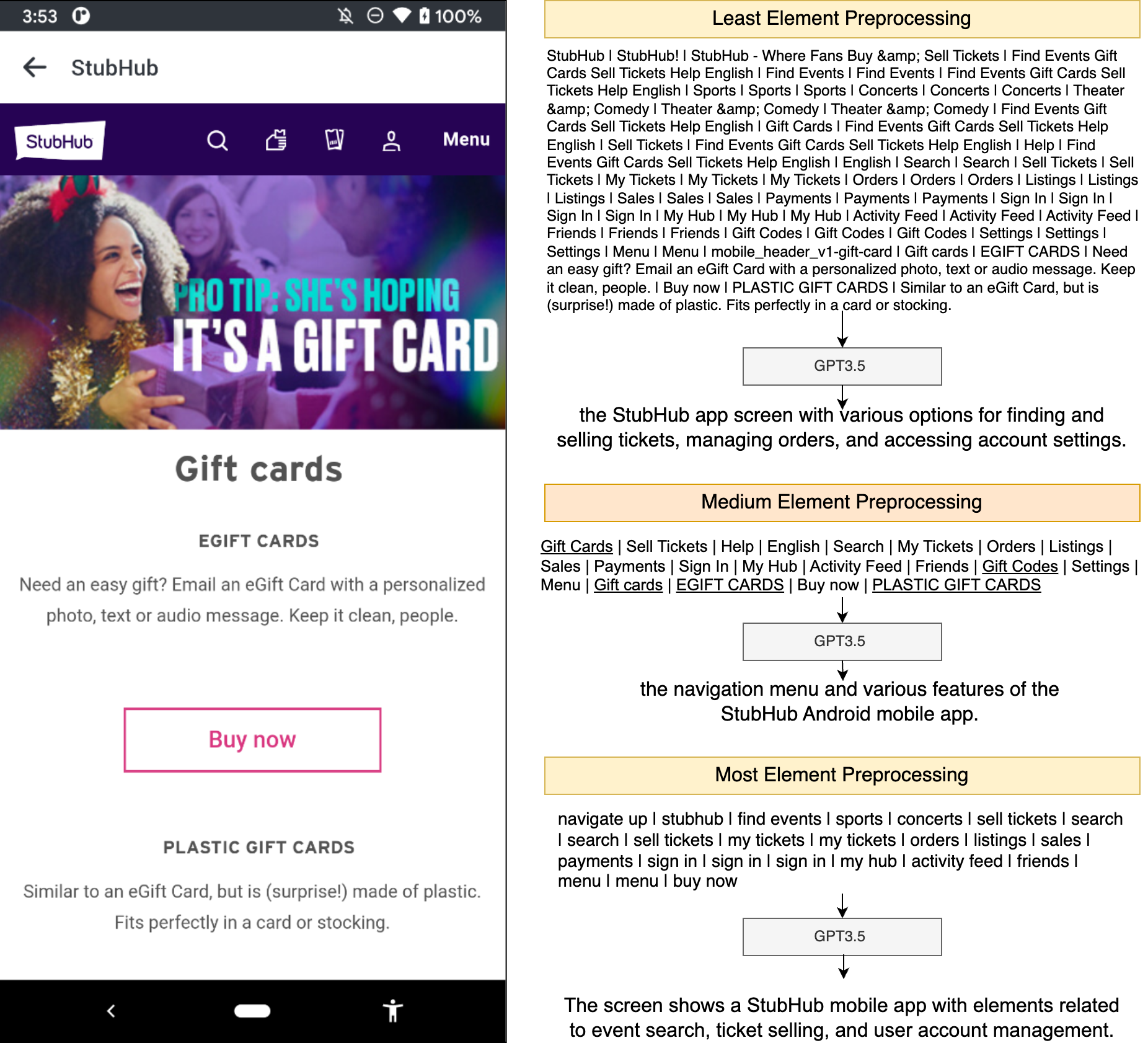}
    \caption{A GPT3.5 Turbo Failure Case. We illustrate a UI screen from the StubHub app that is selling gift cards. Regardless of the degree of UI element text processing, no input is able to make GPT generate a caption which fully captures the purpose of the screen.}
    \label{fig:gptfail}
\end{figure*}
In Figure~\ref{fig:gptfail}, we show an example failure. The StubHub screen concerns E-Gift cards, but none of the input element processing variants we tried were able to correct the focus of the GPT output. We tried several element processing variants which include the most stringent processing (that of Spotlight), the completely raw and unprocessed text, and the in-between that results from our final processing rules.

\section{Downstream UI Tasks}
\label{sec:taskexappendix}
We now provide additional details concerning our downstream benchmark tasks.
\subsection{Finetuning Set Up}
For all tasks other than screen summarization, we input a question $Q$ prompting the model during finetuning. Below, we define the questions for each task:
\begin{align*}
    Q_{widget} &= \text{What describes the functionality of} \\
    & \text{ the UI object found at } [x_1, y_1, x_2, y_2]\text{?''}\\ \\
    Q_{tap} &= \text{``Can the UI object found at }\\
    & [x_1, y_1, x_2, y_2] \text{ be interacted with?''}\\ \\
    Q_{ground} &= \text{``What command refers to the element} \\
    & \text{located at } [x_1, y_1, x_2, y_2]\text{?''}
\end{align*}

Note that for the tappability prediction task, there is a class imbalance (approximately 1:3) of not-tappable to tappable examples. Due to this and the small dataset size, we upsample the not-tappable class by 4x to ensure it is more highly weighted during training and to try to minimize overfitting.

\subsection{Formulating Prediction Tasks as Text Generation}
\label{sec:modelappendix}

We train and evaluate two predictive tasks: tappability prediction and language command grounding. We reformulate both to be possible as text generation tasks, which was also done by Spotlight. For tappability, we have the language model in BLIP-2 decode a caption instead of a class. Specifically, tappable is represented by the answer ``yes the object is interactive,'' while not tappable is represented by ``no the object is not interactive.'' These are answers to the questions posed in the above Appendix section. These captions can then be converted to classes for F1 score and accuracy computation. 

For language command grounding, instead of predicting an element (\ie, predicting which element matches the command) during training, we aim to decode the original complete command given the target element. Then, at test time, we generate instruction captions for all possible elements on the input UI. We perform classification by selecting the element with the instruction caption closest to the ground truth command. If the ground truth element's generated command is the highest scoring, we consider it the prediction. Note that if the score of the target element is equal to the score of other non-target objects, we still consider it a valid prediction (so long as they're the highest).

This process is heavily dependent on the metric used for caption similarity. Due to BLEURT being more highly correlated with human judgement, we use it for computing the similarity between the true language grounding command and the generated element instruction. We also include ablations for which metric was used in Appendix \ref{sec:ablateappendix}.

\section{Computational Details}
We trained BLIP-2 models with 48GB GPU cards (A100, A40, A6000, or L40 NVIDIA cards). Pretraining required 3 days for larger datasets (element list and screen captioning baselines) with 4 GPUs (using multi-GPU training). Training Textual Foresight took half of the time, at around 1.5 days. Finetuning time varies by dataset as well, varying between 2-6 hours for each experiment. We typically use the multi-GPU set up during finetuning as well. We have the same parameter counts as BLIP-2: 188M trainable parameters during pretraining, and 1.2B parameters during finetuning. 

Note that when we make training dataset comparisons to prior work Spotlight, we are considering the training data used for UI representation learning. Both our work and Spotlight initialize models with pretrained checkpoints (ours from pretrained BLIP-2, Spotlight from pretrained T5 and ViT models).

\section{Ablations}
\label{sec:ablateappendix}
We now report results from additional ablations that were run, including more evaluation metrics, results when using OPT in place of the FlanT5 language model, performance with different learning rate and warm up ablations, and results when training from a BLIP-2 checkpoint versus from scratch.

\subsection{Additional Metrics}
We report additional metrics for all downstream UI tasks in Tables~\ref{tab:fortunexpsgenmetricablate} and \ref{tab:fortunexpspredmetricablate}. For screen summarization and element captioning tasks, we additionally report BERTScore and BLEURT text similarity metrics. We use the D-12 distilled version of the latest BLEURT-20 variant due to computational constraints, but found only small differences between the distilled and non-distilled models. Generally BERTScore and BLEURT are less sensitive to changes in captions, but trends are consistent for element captioning, and the metrics do not seem to capture differences for screen summarization.

For tappability prediction, we additionally include accuracy, which holds the same trend as our results with F1 score. For language grounding, we show how the metric we use to determine the best generated instruction command impacts accuracy. While it changes absolute values, the respective trends between methods stay the same.
\begin{table*}[t!]
\centering
\begin{tabular}{l|c|c|c|c|c|c}
\hline
\multirow{3}{*}{\bf Model} &\multicolumn{6}{c}{\bf Task} \\ 
\cline{2-7}
& \multicolumn{3}{c|}{Screen Summarization} & \multicolumn{3}{c}{Element Captioning} \\
\cline{2-7}
&  CIDEr & BERTScore & BLEURT &  CIDEr & BERTScore & BLEURT \\
\hline
Screen2Words & 61.3 & -- & -- & -- & -- & -- \\
Widget Caption &-- & -- & -- & 97.0 & -- & --\\
VUT & 65.6 & -- & -- & 99.3  & -- & --\\
Spotlight & 106.7 & -- & -- & 141.8 & -- & -- \\ 
\hline
BLIP-2 (Original) & 125.1 & 0.90 & 0.65 & 121.4 & 0.88 & 0.47 \\
\hspace{1mm} Screen Caption & 125.7 & 0.90 & 0.65 & 118.9 & 0.88 & 0.46 \\
\hspace{1mm} Element List & 127.9 & 0.90 & 0.65 & 121.6 & 0.88 & 0.47 \\
\hspace{1mm} Textual Foresight & 125.4 & 0.90 & 0.64 & 128.0 & 0.89 & 0.49 \\
 \hline
\end{tabular}
\caption{Finetuning Generative Task Results with Additional Metrics. Prior work includes task specific methods Screen2Words~\citep{screen2words} and Widget Caption~\citep{widgetcap}, multitask model VUT~\citep{vut}, and representation learning approach Spotlight~\citep{spotlight}. All of our baselines and Textual Foresight are built upon BLIP-2~\citep{li2023blip2}. CIDEr, BERTScore, and BLEURT-D12 are reported.}
\label{tab:fortunexpsgenmetricablate}
\end{table*}

\begin{table*}
\setlength{\tabcolsep}{4pt}
\renewcommand{\arraystretch}{0.95}
\centering
\begin{tabular}{l|c|c|c|c|c|c}
\hline
\multirow{3}{*}{\bf Model} &\multicolumn{6}{c}{\bf Task} \\ 
\cline{2-7}
& \multicolumn{2}{c|}{Tappability} & \multicolumn{4}{c}{Language Grounding} \\
\cline{2-7}
& F1 & Acc. &  Acc. & Acc. w/ CIDEr & Acc. w/ BERTScore & Acc. w/ BLEURT\\
\hline
Taperception & 85.5 & -- & -- & -- & -- & -- \\
Swearngin \& Li & 87.9 & -- & -- & -- & -- & --\\
MUG & -- & -- & 58.6 & -- & -- & -- \\
VUT & 88.3 & -- & -- & -- & -- & -- \\
Spotlight & 88.4 & -- & -- & -- & -- & -- \\ 
\hline
BLIP-2 (Original) & 63.9 & 69.3 & -- & 29.3 & 21.7 & 29.8\\
\hspace{1mm} Screen Caption & 68.5 & 75.1 & -- & 32.1 & 26.9 & 34.3 \\
\hspace{1mm} Element List & 67.1 & 74.7 & -- & 35.1 & 29.0 & 38.2 \\
\hspace{1mm} Textual Foresight & 74.2 & 82.3 & -- & 37.1 & 30.9 & 39.5 \\
 \hline
\end{tabular}
\caption{Finetuning Predictive Task Results with Additional Metrics. Prior work includes task specific methods Taperception~\citep{tappability}, Swearngin \& Li~\citep{swearngintap}, MUG~\cite{li2022mug}, multitask model VUT~\citep{vut}, and representation learning approach Spotlight~\citep{spotlight}. All of our baselines and Textual Foresight are built upon BLIP-2~\citep{li2023blip2}. F1 Score and Accuracy (Acc.) is reported for tappability. We also report accuracy when using different text similarity metrics for our language grounding set up with CIDEr, BERTScore, BLEURT-20-D12.}
\label{tab:fortunexpspredmetricablate}
\end{table*}

\subsection{OPT and Learning Rate Ablations}
\label{sec:optablate}
Early on we tried different language models in BLIP-2 and different finetuning learning rates. In Table~\ref{tab:optlr}, we show the ablations ran for screen captioning when finetuning the original BLIP-2 model with warmup steps set to 1000. We vary the initial learning rate and try using the FlanT5, OPT2, and OPT6 LLMs.

\begin{table*}[t!]
\centering
\begin{tabular}{l|c|c|c}
\hline
\multirow{2}{*}{\bf Model} & \multirow{2}{*}{\textbf{LLM}} & \multirow{2}{*}{\textbf{Learning Rate}} & \textbf{Screen Summarization} \\ 
\cline{4-4}
&  & & Validation CIDEr \\
\hline
\multirow{6}{*}{BLIP-2 (Original)} & \multirow{2}{*}{FlanT5} & 1e-5 & 124.4 \\
& & 1e-6 & 120.2 \\
\cline{2-4}
& \multirow{2}{*}{OPT2.7B} & 1e-5 & 122.0 \\
& & 1e-6 & 120.5 \\
\cline{2-4}
& \multirow{2}{*}{OPT6.7B} & 1e-5 & 121.6 \\
& & 1e-6 & 119.9 \\
 \hline
\end{tabular}
\caption{Learning Rate and Language Model Ablations. We varied the LLM used as a part of the BLIP-2 framework, trying FlanT5, OPT2, and OPT6 variants. We also tried different learning rates. Both the language model and learning rate were evaluated on validation performance. We include a subset of the ablations here for the screen summarization task.}
\label{tab:optlr}
\end{table*}

\subsection{Pretrained Checkpoint and Warm Up Ablations}
In Tables~\ref{tab:screenwarmandckpt}-\ref{tab:tapwarmandckpt} we include additional ablations varying the pretrained checkpoint and number of warm up steps during finetuning. We either initialize from a stage one BLIP-2 checkpoint or train the model from scratch. Initializing the model consistently performs better. Then, we try three different values of warm up steps depending on the size of the finetuning dataset: the number of steps for one epoch with our batch size, roughly half of that, and 1000 steps. We include 1k warm up steps because that was the default used for finetuning in the original BLIP-2 model. The best number of warmup step varies by pretrained model.
\begin{table*}[t!]
\centering
\begin{tabular}{l|c|c|c}
\hline
\multirow{2}{*}{\bf Model} & \multirow{2}{*}{\textbf{Initialization}} & \multirow{2}{*}{\textbf{\# Warm Up Steps}} & \textbf{Screen Summarization} \\ 
\cline{4-4}
&  & & Validation CIDEr \\
\hline
\multirow{3}{*}{BLIP-2 (Original)} & \multirow{3}{*}{BLIP-2} &  1000 & 124.4 \\
& & 2500 & 124.8 \\
& & 4919 & 124.6 \\
\hline 
\multirow{6}{*}{Screen Caption} & \multirow{3}{*}{BLIP-2}&  1000  & 125.6 \\
& & 2500 & 124.3 \\
& & 4919 & 125.4\\
\cline{2-4}
& \multirow{3}{*}{Scratch}& 1000 & 120.1 \\
& & 2500 & 117.4 \\
& & 4919 & 119.0 \\
\hline
\multirow{6}{*}{Element List} & \multirow{3}{*}{BLIP-2} & 1000 & 126.9 \\
& & 2500 & 127.4 \\
& & 4919 & 126.4 \\
\cline{2-4}
& \multirow{3}{*}{Scratch} & 1000 & 122.1 \\
& & 2500 & 117.2 \\
& & 4919 & 120.3 \\
\hline
\multirow{6}{*}{Textual Foresight}& \multirow{3}{*}{BLIP-2} & 1000 & 124.1 \\
& & 2500 & 125.0 \\
& & 4919 & 125.9 \\
\cline{2-4}
& \multirow{3}{*}{Scratch} & 1000 &  109.1 \\
& & 2500 & 108.6 \\
& & 4919 & 108.1 \\
 \hline
\end{tabular}
\caption{Screen Summarization Pretrained Checkpoint and Warmup Ablations. We varied whether the pretrained model was initialized with or without a BLIP-2 checkpoint. For each task, we also parameter tune the number of warm up steps and select the best model based on validation performance.}
\label{tab:screenwarmandckpt}
\end{table*}

\begin{table*}[t!]
\centering
\begin{tabular}{l|c|c|c}
\hline
\multirow{2}{*}{\bf Model} & \multirow{2}{*}{\textbf{Initialization}} & \multirow{2}{*}{\textbf{\# Warm Up Steps}} & \textbf{Element Captioning} \\ 
\cline{4-4}
&  & & Validation CIDEr \\
\hline
\multirow{3}{*}{BLIP-2 (Original)} & \multirow{3}{*}{BLIP-2} & 1000 & 123.6 \\
& & 3500 & 124.5 \\
& & 6835 & 121.6 \\
\hline 
\multirow{6}{*}{Screen Caption} & \multirow{3}{*}{BLIP-2} & 1000 & 122.6\\
& & 3500 & 124.3  \\
& & 6835 & 121.4\\
\cline{2-4}
& \multirow{3}{*}{Scratch} & 1000 & 112.2 \\
& & 3500 & 109.5 \\
& & 6835 & 110.3 \\
\hline
\multirow{6}{*}{Element List} & \multirow{3}{*}{BLIP-2} & 1000 & 126.9 \\
& & 3500 & 126.5 \\
& & 6835 & 123.9 \\
\cline{2-4}
& \multirow{3}{*}{Scratch} & 1000 & 127.6 \\
& & 3500 & 125.9 \\
& & 6835 & 126.1 \\
\hline
\multirow{6}{*}{Textual Foresight}& \multirow{3}{*}{BLIP-2} & 1000 & 133.3 \\
& & 3500 & 132.7 \\
& & 6835 & 131.4 \\
\cline{2-4}
& \multirow{3}{*}{Scratch} & 1000 & 119.1\\
& & 3500 & 117.6 \\
& & 6835 & 117.3 \\
 \hline
\end{tabular}
\caption{Element Captioning Pretrained Checkpoint and Warmup Ablations. We varied whether the pretrained model was initialized with or without a BLIP-2 checkpoint. For each task, we also parameter tune the number of warm up steps and select the best model based on validation performance.}
\label{tab:elemwarmandckpt}
\end{table*}

\begin{table*}[t!]
\centering
\begin{tabular}{l|c|c|c}
\hline
\multirow{3}{*}{\bf Model} & \multirow{3}{*}{\textbf{Initialization}} & \multirow{2}{*}{\textbf{\# Warm Up Steps}} & \textbf{Tappability Prediction} \\ 
\cline{4-4}
&  & & Validation F1 \\
\hline
\multirow{3}{*}{BLIP-2 (Original)} & \multirow{3}{*}{BLIP-2} & 500 & 64.5 \\
& & 1000 & 59.9 \\
& & 1124 & 66.1 \\
\hline 
\multirow{6}{*}{Screen Caption} & \multirow{3}{*}{BLIP-2} & 500 & 64.9 \\
& & 1000 & 63.4\\
& & 1124 & 51.0 \\
\cline{2-4}
& \multirow{3}{*}{Scratch} & 500 & 68.5\\
& & 1000 & 69.2\\
& & 1124 & 67.9 \\

\hline
\multirow{6}{*}{Element List} & \multirow{3}{*}{BLIP-2} & 500 & 63.2 \\
& & 1000 & 59.4 \\
& & 1124 &  65.8\\
\cline{2-4}
& \multirow{3}{*}{Scratch} & 500 & 68.2 \\
& & 1000 & 69.6 \\
& & 1124 & 68.6 \\
\hline
\multirow{6}{*}{Textual Foresight}& \multirow{3}{*}{BLIP-2} & 500 & 73.3 \\
& & 1000 & 69.9 \\
& & 1124 & 74.4 \\
\cline{2-4}
& \multirow{3}{*}{Scratch} & 500 & 69.0 \\
& & 1000 & 69.3 \\
& & 1124 & 69.0 \\
 \hline
\end{tabular}
\caption{Tappability Prediction Pretrained Checkpoint and Warmup Ablations. We varied whether the pretrained model was initialized with or without a BLIP-2 checkpoint. For each task, we also parameter tune the number of warm up steps and select the best model based on validation performance.}
\label{tab:tapwarmandckpt}
\end{table*}


\end{document}